\documentclass[twoside,11pt]{article}

\usepackage{jmlr2e}

\usepackage{amsmath}
\usepackage{amssymb}
\usepackage{amsthm}
\usepackage{graphicx}
\usepackage{color}
\usepackage{mathrsfs}

\usepackage{epsfig}
\usepackage{natbib}

\newcommand{\be}{\begin{equation}}
\newcommand{\ee}{\end{equation}}
\newcommand{\bes}{\begin{equation*}}
\newcommand{\ees}{\end{equation*}}
\newcommand{\beqn}{\begin{eqnarray}}
\newcommand{\eeqn}{\end{eqnarray}}
\newcommand{\beqns}{\begin{eqnarray*}}
\newcommand{\eeqns}{\end{eqnarray*}}

\newcommand{\lkr}{\left(}
\newcommand{\lkv}{\left[}
\newcommand{\rkv}{\right]}
\newcommand{\rkr}{\right)}
\newcommand{\lfi}{\left\{}
\newcommand{\rfi}{\right\}}
\newcommand{\di}{\displaystyle}

\long\def\ignore#1{}

\newcommand{\bm}[1]{{\mbox{\mathversion{bold}$#1$}}}

\newcommand{\R}{\mathbb{R}}

\newcommand{\del}{\delta}

\newcommand{\Del}{\Delta}

\newcommand{\ga}{\gamma}
\newcommand{\gam}{\gamma}

\newcommand{\om}{\omega}
\newcommand{\lam}{\lambda}

\newcommand{\sig}{\sigma}

\newcommand{\Lam}{\Lambda}
\newcommand{\Om}{\Omega}

\newcommand{\EE}{\ensuremath{{\mathbb E}}}

\newcommand{\PP}{\ensuremath{{\mathbb P}}}

\newcommand{\RR}{\ensuremath{{\mathbb R}}}

\newcommand{\diag}{\mbox{diag}}

\newcommand{\Tr}{\mbox{Tr}}

\newcommand{\rank}{\mbox{rank}}

\newcommand{\vect}{\mbox{vec}}
\newcommand{\Pen}{\mbox{Pen}}
\newcommand{\Err}{\mbox{Err}}
 \newcommand{\su}{\text{supp}}

\newtheorem{thm}{Theorem}
\newtheorem{lem}{Lemma}

\newtheorem{rem}{Remark}
\newtheorem{ex}{Example}

\newcommand{\brJ}{\breve{J}}

\newcommand{\calM}{{\mathcal{M}}}

\newcommand{\calN}{{\cal N}}

\newcommand{\scrP}{\mathscr{P}}

\newcommand{\scrPZK}{\mathscr{P}_{Z,K}}
\newcommand{\scrPhZK}{\mathscr{P}_{\hat{Z}_K,K}}

\newcommand{\norm} [1]{\left\|#1 \right\|}

\newcommand{\pijbAZ} { \Pi_{J^{(k,l)}} (A^{(k,l)} (Z,K))}
\newcommand{\pijhbAZh} {\Pi_{\hat{J}^{(k,l)}}  \lkr A^{(k,l)} (\hat{Z}, \hat{K})\rkr}
\newcommand{\pijhbPsZh} {\Pi_{\hat{J}^{(k,l)}} ({P_{*}}^{(k,l)} (\hat{Z}, \hat{K}))}

\newcommand{\pijkl}{\Pi_{J^{(k,l)}}}

\newcommand{\pibrjkl}{\Pi_{\breve{J}^{(k,l)}}}

\newcommand{\pijhPsZh} {\Pi_{\hat{J}} ({P_{*}} (\hat{Z}, \hat{K})) }
\newcommand{\pijhEZh} {\Pi_{\hat{J} } (\Xi (\hat{Z}, \hat{K})) }

\newcommand{\cC}{\check{C}}
\newcommand{\tC}{\tilde{C}}

\newcommand{\tJk}{\tilde{J}_k}

\newcommand{\cJk}{\check{J}_k}

\newcommand{\tNk}{\tilde{N}_k}

\newcommand{\cNk}{\check{N}_k}
\newcommand{\cNl}{\check{N}_l}
\newcommand{\tbe}{\tilde{\beta}}
\newcommand{\cbe}{\check{\beta}}
\newcommand{\tbek}{\tilde{\beta}_k}

\newcommand{\cbek}{\check{\beta}_k}

\newcommand{\tbk}{\tilde{B}_k}

\newcommand{\cbl}{\check{B}_l}
\newcommand{\tbl}{\tilde{B}_l}
\newcommand{\tlamkk}{\tilde{\Lam}^{(k,k)}}
 
\newcommand{\tlamkl}{\tilde{\Lam}^{(k,l)}}

\newcommand{\clamkk}{\check{\Lam}^{(k,k)}}
\newcommand{\clamll}{\check{\Lam}^{(l,l)}} 
\newcommand{\clamkl}{\check{\Lam}^{(k,l)}}
\newcommand{\clamlk}{\check{\Lam}^{(l,k)}}

\newcommand{\tlam}{\tilde{\lam}}

\newcommand{\tPkk}{\tilde{P}^{(k,k)}}
\newcommand{\tPo}{\tilde{P}^{(1,1)}}
\newcommand{\tPt}{\tilde{P}^{(2,2)}}
\newcommand{\tPot}{\tilde{P}^{(1,2)}}

\newcommand{\tHk}{\tilde{H}_k}
\newcommand{\tH}{\tilde{H}}

\firstpageno{1}

\begin{document}

\title{Sparse Popularity Adjusted Stochastic Block Model}

\author{\name Majid Noroozi \email majid.noroozi@wustl.edu \\
       \addr Department of Mathematics and Statistics\\
       Washington University in St. Louis\\
     St. Louis, MO 63130, USA
      \AND
     \name Marianna Pensky\footnote{Corresponding Author} \email Marianna.Pensky@ucf.edu \\
      \addr Department of Mathematics\\
       University of Central Florida\\
     Orlando, FL 32816, USA     
       \AND
    \name Ramchandra Rimal \email ramchandra.rimal@mtsu.edu \\
      \addr Department of Mathematical Sciences\\
      Middle Tennessee State University\\
     Murfreesboro, TN 37132, USA      }

\editor{}

\maketitle

\begin{abstract}
In the present paper we study a sparse stochastic network enabled with a block structure.
The popular Stochastic Block Model (SBM) and the Degree Corrected Block Model (DCBM)
address  sparsity by placing an upper bound on the maximum probability of connections 
between any pair of nodes. As a result, sparsity describes only the behavior of network as a whole, 
without distinguishing between the block-dependent sparsity patterns. 
To the best of our knowledge, the recently introduced Popularity Adjusted  Block Model (PABM)
is the only block model that allows to introduce a {\it structural sparsity} where some
probabilities of connections are identically equal to zero while the rest of them remain 
above a certain threshold. The latter presents a more nuanced view of the network. 
%
\end{abstract}


\section{Introduction}
\label{sec:intro}


\subsection{Stochastic Block Models}
\label{sec:block}
 
The  last few years   have seen a surge of interest in stochastic  network models.
Indeed, such models appear in a variety of applications ranging from social to biological sciences. 
Stochastic networks can be described in a variety of ways, however, in the last decade stochastic block models 
attracted more and more attention due to their ability to
summarize data in a compact and intuitive way and to uncover   low-dimensional structures that 
fully describe a given network.

In this paper, we consider an undirected network with $n$ nodes and no self-loops and multiple edges. 
Let $A \in \{0,1\}^{n\times n}$ be the symmetric adjacency  matrix of the network with
$A_{i,j}=1$ if there is a connection between nodes $i$ and $j$,  and $A_{i,j}=0$
otherwise. We assume that 
\be \label{eq:A_Bern_P}
 A_{i,j}\sim \mbox{Bernoulli}(P_{i,j}),  \quad 1 \leq i \leq j \leq n,
\ee
where  $A_{i,j}$  are conditionally independent 
given $P_{i,j}$ and  $A_{i,j} = A_{j,i}$, $P_{i,j} = P_{j,i}$ for $i>j$.

The block models assume that each node  in the network belongs to one of $K$ distinct blocks or communities $\calN_k$, $k=1, \cdots, K$. 
The communities are described by the vector $c$ of community assignment,  with $c_i = k$ if the   node $i$ belongs to the   community $k$.
One can also consider a corresponding {\it membership} (or {\it clustering}) matrix $Z \in \{0,1\}^{n\times K} $     
such that $Z_{i,k}=1$ iff $i \in \calN_k$, $i=1, \ldots, n$. The degree of a node $i$ and its expected degree are defined, respectively, as 
the number of edges and the sum of probabilities of connections between the node $i$ and the rest of the nodes.

One of the features of the block models is that they assume that  the probability of connection between node 
$i \in \calN_k$ and node $j \in  \calN_l$  depends on the pair of blocks $(k,l)$ to which nodes $(i,j)$ belong.
In particular, the Stochastic Block Model (SBM) assumes that  the probability of connection between nodes 
is completely defined by the communities to which they belong, so that, for any pair of nodes  $(i,j)$,
one has $P_{i,j} = B_{c_i,c_j}$  where $B_{k,l}$ is the probability of connection between communities $k$ and $l$.
In particular, under the SBM, all nodes from the same community have the same expected degree.

Since the real life networks usually contain a very small number of high-degree nodes while the rest 
of the nodes have very few connections (low degree), the SBM model fails to explain the structure of 
many networks that occur  in practice. 
The  Degree Corrected Block Model (DCBM) addresses this deficiency by  allowing these probabilities 
to be multiplied by the node-dependent weights (see, e.g., \cite{chen2018}, \cite{Karrer2011StochasticBA}, 
\cite{zhao2012consistency}  among others). Under the DCBM, the elements of matrix $P$ are modeled as 
$P_{i,j }= \theta_i\, B_{c_i,c_j}\theta_j$, where $\theta_i$, $i=1,\ldots,n$, are the degree parameters of the   nodes, 
and $B$ is the $(K \times K)$  matrix of baseline interaction between communities. 
Identifiability of the parameters is usually ensured by a constraint of the form  
$\sum_{i\in \mathcal{N}_k} \theta_i = 1$ for all    $k = 1,\ldots,K$
(see, e.g., \cite{Karrer2011StochasticBA}).

The Popularity Adjusted Block Model (PABM), introduced by \cite{RePEc:bla:jorssb:v:80:y:2018:i:2:p:365-386} 
and subsequently studied in \cite{https://doi.org/10.1111/rssb.12410}, provides a generalization of both the SBM 
and the DCBM. The DCBM enables a more flexible spectral structure of matrix $P$ which is especially useful 
in the cases when the mixed membership models cannot be employed.  We are particularly interested in the PABM since,
to the best of our knowledge, it is the only block model that allows to model structural sparsity 
in the connections between the nodes in the network.

In order to understand the PABM, consider a rearranged version $P(Z,K)$ of matrix $P$ where its  first 
$n_{1}$ rows correspond to  nodes from class 1, the next $n_{2}$ rows correspond to  nodes from class 2 
and the last $n_K$ rows correspond to  nodes from class $K$. Denote the $(k,l)$-th block of matrix $P(Z,K)$ by $P^{(k,l)} (Z,K)$. 
Then, sub-matrix $P^{(k,l)} (Z,K) \in [0,1]^{n_k \times n_l}$ corresponds to pairs of nodes in communities $(k,l)$ respectively.
It is easy to see that in the SBM,   $P^{(k,l)} (Z,K)$  has all elements equal to   $B_{k,l}$, 
while in the DCBM, $P^{(k,l)}  = B_{k,l} \theta^{(k)} (\theta^{(l)})^T$
where $\theta^{(k)}$ is the sub-vector of vector $\theta$ that contains weights for the nodes in community $k$. 
Under the   PABM,  each pair of blocks $P^{(k,l)} (Z,K)$ and $P^{(l,k)} (Z,K)$  is defined using a  unique combination 
of vectors $\Lambda^{(l,k)}$  as follows: 
\be \label{eq:block_structure}
P^{(k,l)} (Z, K) =   [P^{(l,k)} (Z,K)]^T = \Lambda^{(k,l)} \, [\Lambda^{(l,k)}]^{T} \in [0,1]^{ n_k \times n_l}, \quad k,l=1, \ldots, K.
\ee 
Here, vectors $\Lambda^{(k,l)} \in [0,1]^{n_k}$, $k=1, \ldots, K$, form 
  column $l$ of matrix  $\Lambda \in [0,1]^{n \times K}$  given by
\be \label{eq:Lambda}
\Lambda=
  \begin{bmatrix}
    \Lambda^{(1,1)} & \Lambda^{(1,2)} & \cdots  & \Lambda^{(1,K)}\\
    \Lambda^{(2,1)} & \Lambda^{(2,2)} & \cdots  & \Lambda^{(2,K)}  \\
    \vdots & \vdots& \cdots& \vdots\\
    \Lambda^{(K,1)} & \Lambda^{(K,2)} & \cdots  & \Lambda^{(K,K)} \\
  \end{bmatrix}
\ee 
Vector $\Lambda^{(k,l)}$ represents the {\it popularity} (or, the level of interaction)  of  nodes in class $k$
with respect to class $l$. The PABM allows higher degree of flexibility in modeling the probability matrix and, in addition, does 
not require any identifiability conditions for its fitting, thus, providing an attractive alternative to SBM  and  DCBM.


\subsection{Sparsity in Block Models}
\label{sec:sparsity}

The real life networks are usually sparse in a sense that a large number of nodes have small degrees.
One of the shortcomings of both the SBM and the DCBM is that they do not allow to efficiently model sparsity. 
 
Specifically, in majority of high-dimensional setting, ``sparsity'' means  {\it structural sparsity} and 
establishes that some parameters of the model are equal to zero and have no effect on the variables of interest.
Finding  the set of nonzero parameters in such models is one of the goals of the inference.
This is true in, for example, high-dimensional regression model where identification of the set of nonzero coefficients
is crucial for understanding which independent variables affect the variable of interest. However, 
the traditional stochastic block models do not allow to model sparsity in a structural way. 
The latter is due to simplistic modeling of connection probabilities.

Indeed, for the SBM, it is not realistic to assume that all nodes in a pair of communities have no connections, hence,
in the SBM setting, one does not assume that the  block probabilities  $B_{k,l}=0$  for some   $k$ and $l$. 
The DCBM is not very different in this respect, since setting any node-specific weight to zero  will 
force the respective node to be totally disconnected from the network. 
For this reason, unlike in other numerous statistical settings,   sparsity in block models is defined as a 
low maximum probability of connections between the nodes: $\di \max_{i,j} P_{i,j} \leq \tau(n)$ where $\tau(n)\  \to \ 0$
as $n \to \infty$    (see, e.g., \cite{klopp2017}, \cite{lei2015}). 
As a result, sparsity describes only the behavior of network as a whole, without distinguishing between the 
block-dependent sparsity patterns. In addition,  the above definition of sparsity has  other drawbacks. 
In particular, one has to estimate {\it every} probability of connections  $B_{k,l}$, no matter how small it is, 
and, in many settings (see, e.g., \cite{klopp2017}),  in order to take  advantage of the fact that $P_{i,j}$ 
are bounded above by $\tau(n)$, one needs to incorporate this unknown value into the estimation process.

To the best of our knowledge, the PABM is the only existing block model that allows to 
model sparsity as {\it structural sparsity} where some connection probabilities are equal to zero, while
the average connection probabilities between classes are above certain level, and the network is connected. 
In the context of PABM, setting $\Lambda^{(k,l)}_i = 0$
simply means that that node $i$ in class $k$ is not active  (``popular") in class $l$. 
This, nevertheless,  does not 
prevent this node from having high probability of connection with nodes in another class.  
 Setting some   elements of vectors $\Lambda^{(k,l)}$ to  zero   will merely lead to
some of the rows (columns) of sub-matrices $P^{(k,l)} (Z,K)$ being zero. 
Moreover, since $A_{i,j}$ are Bernoulli variables with the means $P_{i,j}$, 
those zeros are fairly easy to identify, as $P_{i,j}=0$  implies $A_{i,j}=0$.

Identification of the set of zeros in the sub-columns $\Lambda^{(k,l)}$ of  matrix  $\Lambda$
gives the nuanced picture of the behavioral patterns of the nodes in the network and 
leads to  a better understanding of network topology. Moreover, it allows to improve the precision of 
estimation of the matrix of connection probabilities, since it is well known that, 
when many of the elements of a vector or a matrix are identical zeros, 
identifying those zeros and estimating the rest of the elements leads 
to a smaller error than when this information is ignored.

In summary, to the best of our knowledge, our paper is the  first paper that studies 
structural sparsity in stochastic block models and the PABM is the only block model that allows the treatment.

The rest of the paper is organized as follows.
Section~\ref{sec:est_clust} is the key part of the paper. After introducing notations in Section~\ref{sec:notation},  
we review the PABM  and convey the structure of the probability matrix  in Section~\ref{sec:structure}. 
Section~\ref{sec:opt_proc} formulates an optimization procedure for estimation and clustering. 
Furthermore,  Section~\ref{sec:support} suggests two possible expressions for the penalties and 
examines the support sets of the true and estimated probability matrices. 
Section~\ref{sec:errors} produces upper bounds on the estimation and clustering errors.
Since the optimization procedure in Section~\ref{sec:opt_proc} is NP-hard, Section~\ref{sec:clustering}   
discusses implementation of the community detection via sparse subspace clustering.
Sections~\ref{sec:simulations}~and~\ref{sec:real_data}   complement the theory with simulations on synthetic networks
and real data examples.  Finally, Appendix~\ref{sec:appendA} presents simulation results for the precision 
of estimation of the number of communities, and also contains the proofs of the statements in the paper.


\section{Estimation and  Clustering in Sparse PABM}
\label{sec:est_clust}
 

\subsection{Notation}
\label{sec:notation}

For any two positive sequences $\{ a_n\}$ and $\{ b_n\}$, $a_n \asymp b_n$ means that 
there exists a  constant  $C>0$ independent of $n$ such that $C^{-1} a_n \leq b_n \leq C a_n$
for any $n$. For any set $\Om$, denote cardinality of $\Om$ by $|\Om|$.
For any numbers $a$ and $b$, $a \wedge b = \min (a,b)$.
%
For any vector $ t \in \RR^p$, denote  its $\ell_2$, $\ell_1$, $\ell_0$ and $\ell_\infty$ norms by, 
respectively,  $\|  t\|$, $\|  t\|_1$,  $\|  t\|_0$ and $\|  t\|_\infty$.
Denote by $1_m$  the $m$-dimensional column vector with all components equal to one.  
For any matrix $A$,  denote its spectral and Frobenius norms by, respectively,  $\|  A \|_{op}$ and $\|  A \|_F$.
Let $\vect(A)$ be the vector obtained from matrix $A$ by sequentially stacking its columns.  
Denote column $i$ of matrix $A$ by $A_{:,i}$.

Denote by $\Pi_{J}(X)$, the projection of a matrix $X: n\times m$ onto the set of matrices with nonzero elements in the set 
$J = J_1 \times J_2 =  \{(i,j): i \in J_1,\ j \in J_2 \}$.
Denote by $\Pi_{(1)}(X)$ the best rank one approximation of matrix $X$ and by $\Pi_{u,v}(X)$ the rank one projection of $X$
onto pair of unit vectors $u,v$ given by 
\be \label{rank1_Proj}
\Pi_{u,v}(X) = (uu^T)X(vv^T).  
\ee  
Then, $\Pi_{(1)}(X) = \Pi_{u,v}(X)$ provided $(u,v)$ is a pair of singular vectors of $X$ corresponding to the largest singular value.

Denote by $\calM_{n,K}$ a collection of   clustering  matrices $Z \in \{0,1\}^{n\times K} $  
 such that $Z_{i,k}=1$ iff $i \in \calN_k$, $i=1, \ldots, n$, and 
$Z^T Z = \diag (n_1, \ldots, n_K)$ where  $n_k = |\calN_k|$ is the size of community $k$, where $k=1, \ldots, K$.
Denote by  $\scrPZK \in \{0,1\}^{n \times n}$  the   permutation matrix corresponding to $Z \in \calM_{n,K}$ that rearranges  
any matrix   $B \in \RR^{n \times n}$, so that its  first $n_{1}$ rows correspond to  nodes from class 1, 
the next $n_{2}$ rows correspond to  nodes from class 2 and the last $n_K$ rows correspond to  nodes from class $K$. 
Recall that $\scrPZK$ is an orthogonal matrix with $\scrPZK^{-1} = \scrPZK^T$. 
For any $\scrPZK$ and any matrix $B \in \RR^{n \times n}$ denote the permuted matrix and its blocks by, respectively,
$B(Z,K)$ and $B^{(k,l)} (Z,K)$, where  $B^{(k,l)}(Z,K) \in \RR^{n_k \times n_l}$,  $k,l=1, \ldots, K$,  and
\be  \label{eq:permute}
B(Z,K) = \scrPZK^T B \scrPZK,  \quad   \quad B = \scrPZK  B(Z,K) \scrPZK^T.
\ee  
Also, throughout the paper, we use the star symbol to identify the true quantities. In particular, we 
denote the true matrix of connection probabilities by $P_*$ 
and the true clustering matrix that partitions $n$ nodes into $K_*$ communities by $Z_*$.


\subsection{The Structural Sparsity of the Probability Matrix }
\label{sec:structure}

Consider the problem of estimation and clustering of the true matrix  $P_*$ of the probabilities of the connection between the nodes.
Consider a block $P_*^{(k,l)}(Z_*, K_*)$ of the rearranged version $P_*(Z_*, K_*)$ of $P_*$. 
Let $\Lam_* \equiv \Lam (Z_*, K_*) \in [0,1]^{n \times K_*}$ be a  block matrix  with each column $l$ partitioned into $K_*$
blocks $\Lam_*^{(k,l)} \equiv  \Lam_*^{(k,l)}(Z_*, K_*)$. 
Here, $\Lambda_*^{(k,l)} \in [0,1]^{n_k}$  and $\Lambda_*^{(l,k)} \in [0,1]^{n_l}$ are the column vectors and 
$P_{*}^{(k,l)} (Z_{*}, K_*)$ follows \eqref{eq:block_structure}, i.e., 
$P_{*}^{(k,l)} (Z_{*}, K_*) =     \Lambda_*^{(k,l)} \, [\Lambda_*^{(l,k)}]^{T}$.
Hence,  $P_*^{(k,l)} (Z_*, K_*)$ are rank-one matrices such that $P_*^{(k,l)} (Z_*, K_*) = [P_*^{(l,k)} (Z_*, K_*)]^T$ 
and that each pair of blocks $P_*^{(k,l)}$ and $P_*^{(l,k)}$, involves a  unique combination of vectors 
$\Lambda_*^{(k,l)}$ and $\Lambda_*^{(l,k)}$, $k,l=1, \ldots, K_*$.

Vectors $\Lambda_*^{(k,l)}$ and $\Lambda_*^{(l,k)}$ describe the heterogeneity of the connections of nodes in the pair of communities $(k,l)$. 
While, on the average, those communities can be connected, some nodes  in  community $k$ may have no interaction with nodes in community $l$
or vice versa, so that some of the elements of vectors  $\Lambda_*^{(k,l)}$ and $\Lambda_*^{(l,k)}$ can be identical zeros. 
Denote    the set of indices of all nonzero elements of matrix $\Lambda_*$ by 
\bes
J_* \equiv J_* (Z_*, K_*)= \di\bigcup_{k,l=1}^{K} (J_*)_{k,l}.
\ees
Let
\be \label{eq:Jstar}
(J_*)_{k,l} \equiv (J_*)_{k,l} (Z_*, K_*) = \{ i:\ (\Lambda_*)_i^{(k,l)} \neq 0 \}, \quad J_*^{(k,l)}  = (J_*)_{k,l}  \times (J_*)_{l,k}, 
\ee 
be, respectively,  the true support of vector $\Lambda_*^{(k,l)}$ and the set of all ordered pairs of indices (positions) of non-zero elements of 
sub-matrix $P_*^{(k,l)} (Z_*, K_*)$. Here, the elements of $(J_*)_{k,l}$ are enumerated by their corresponding rows in matrix $\Lambda_{*}$.   Then, 
\bes
(P_*)^{(k,l)}_{i,j} (Z_*, K_*) >0 \quad \mbox{iff}\quad (i,j) \in J_*^{(k,l)}
\ees
and row $i$ and column $j$ of  $P_*^{(k,l)} (Z_*, K_*)$ are equal to zero if $i \notin (J_*)_{k,l}$ or  $j \notin (J_*)_{l,k}$.

Note that  the set $J_* \equiv J_* (Z_*, K_*)$   relies upon  the true clustering defined by $K_*$ and $Z_*$.
One can also consider sparsity sets $(\brJ_*)_{k,l} \equiv (\brJ_*)_{k,l} (Z,K)$ and $\brJ_{k,l} \equiv  \brJ_{k,l} (Z,K)$
for an arbitrary $K$ and  matrix $Z \in \calM_{n,K}$ 
\beqn 
(\brJ_*)_{k,l} &  = & \{ i:\,   (P_*)^{(k,l)}_{i,j}(Z,K) \neq 0, \   \mbox{ for some}\   j = 1, \ldots, n_l \}, \nonumber \\ 
&    &     \label{eq:brJ}\\
\brJ_{k,l}  &  = &  \{ i:\,  A^{(k,l)}_{i,j}(Z,K) \neq 0, \  \mbox{ for some}\   j = 1, \ldots, n_l  \}, \nonumber
\eeqn
 where the elements of $(\brJ_*)_{k,l}$ and $\brJ_{k,l}$ are enumerated by their corresponding rows in matrices $P_{*}$ and $A$, respectively.
Examples of the sets $(J_*)_{k,l}$,  $(J_*)^{(k,l)}$, $(\brJ_*)_{k,l}$ and $(\brJ_*)^{(k,l)}$ are considered in Section~\ref{sec:support}.
For any sparsity sets $J_{k,l} \equiv J_{k,l} (Z,K)$,  define, similarly to \eqref{eq:Jstar}, 
\beqn \label{eq:J_total}
J =  \bigcup_{k,l=1}^{K} J_{k,l} \quad \mbox{with} \quad 
J^{(k,l)}  = J_{k,l}  \times J_{l,k}
\eeqn
It follows from the definitions \eqref{eq:brJ} and \eqref{eq:J_total} that, for any $K$, $Z \in \calM_{n,K}$ and
$k,l = 1, \ldots, K$
\be \label{eq:brJstarZK}
\brJ_{k,l} (Z,K) \subseteq (\brJ_*)_{k,l} (Z,K)  \quad \mbox{and} \quad
\brJ (Z,K) \subseteq \brJ_* (Z,K).
\ee


\subsection{Optimization Procedure for Estimation and Clustering}
\label{sec:opt_proc}

Observe that although matrices $P_*^{(k,l)} (Z_*, K_*)$   and the sets  $J_*^{(k,l)}$ are well defined, 
vectors $\Lambda_*^{(k,l)}$ and $\Lambda_*^{(l,k)}$ can be determined only up to a multiplicative constant.
In order to avoid this ambiguity, we denote $\Theta_*^{(k,l)} =   \Lambda_*^{(k,l)} [\Lambda_*^{(l,k)}]^T$ 
and recover  matrix $\Theta_*$ with the uniquely defined rank one  blocks $\Theta_*^{(k,l)}$ and their 
supports $J_*^{(k,l)}$, $k,l=1, \ldots, K_*$. 
For this purpose, we need  to solve the following  optimization problem 
\begin{align} \label{eq:opt_main}
   (\hat{\Theta}, \hat{Z},\hat{J}, \hat{K})& \in  \underset{\Theta, Z,J,K}{\text{argmin}} 
\left\{\displaystyle \sum_{k,l = 1}^K \norm {A^{(k,l)}(Z,K) - \Theta^{(k,l)} {(Z,J,K)} }_{F}^2  + \Pen(n,J,K) \right\} \\
 &   \text{s.t.}\quad A(Z,K) = \scrPZK^T A \scrPZK,\  Z \in \calM_{n,K}, \nonumber\\
 &   \su(\Theta^{(k,l)})  = J^{(k,l)} = J_{k,l} \times J_{l,k}, \ 
   \text{rank} (\Theta^{(k,l)}) =1, \  k,l = 1,\ldots,K.\nonumber
 \end{align}
Here, $\hat{\Theta}$ is the block matrix with blocks $ \hat\Theta^{(k,l)}$, $k,l=1, \ldots, K$.

Observe that, if $\hat{Z}$, $\hat{J}$  and $\hat{K}$ were known, the best solution of problem \eqref{eq:opt_main} would be given by 
the best rank one approximations   $\hat{\Theta}^{(k,l)}$ of matrices $A^{(k,l)}(\hat{Z},\hat{K})$, restricted to the sets $\hat{J}^{(k,l)}$
of indices of nonzero elements:
\be \label{eq:Theta_est}
\hat{\Theta}^{(k,l)}  {(\hat{Z}, \hat{J}, \hat{K})} =\Pi_{(1)} \left( \pijhbAZh \right),
\ee 
where  $\Pi_{J^{(k,l)}}\left(A^{(k,l)}\right)$ is  the projection of  matrix $A^{(k,l)}$ onto the set of matrices with the support 
 $J^{(k,l)}$, and $\Pi_{(1)}$ is the best rank one approximation of a matrix. 
Plugging \eqref{eq:Theta_est} into \eqref{eq:opt_main}, we rewrite optimization problem 
\eqref{eq:opt_main} as 
\small{\begin{align}   
 (\hat{Z}, \hat{J}, \hat{K}) & \in  \underset{Z,J,K}{\text{argmin}}  \left\{\displaystyle  
\sum_{k,l = 1}^K \|A^{(k,l)}(Z,K) -  
\Pi_{(1)}  [\pijbAZ] \|_{F}^2  + \Pen(n,J,K)  \right\} \label{eq:opt_ZK}  \\
 & \text{s.t.}\quad A(Z,K) = \scrPZK^T A \scrPZK, \quad Z \in \calM_{n,K},  \nonumber \\
 &   J^{(k,l)} \equiv J^{(k,l)} (Z,K)  = J_{{k,l}} (Z,K) \times J_{{l,k}}  (Z,K), \  k,l = 1,\ldots,K. \nonumber 
\end{align}
}
In practice, in order to obtain $(\hat{Z},\hat{J}, \hat{K})$, one needs to solve optimization problem \eqref{eq:opt_ZK}   for every $K$, 
obtaining
\begin{align}  
(\hat{Z}_K , \hat{J}_K) & \in   \underset{Z, J}{\text{argmin}}  
\left\{\displaystyle \sum_{k,l = 1}^K \norm {A^{(k,l)}(Z,K) -  \Pi_{(1)}\left(\pijbAZ\right)}_{F}^2  + \Pen(n,J,K) \right\}  \label{eq:opt_ZK3} \\
 & \text{s.t.}\quad A(Z,K) = \scrPZK^T A \scrPZK, \quad Z_K \in \calM_{n,K},\nonumber  \\
 &  J^{(k,l)} \equiv J^{(k,l)} (Z,K)  = J_{{k,l}} (Z,K) \times J_{{l,k}} (Z,K),\  k,l = 1,\ldots,K. \nonumber 
\end{align}
and then find $\hat{K}$ as 
\be \label{eq:opt_for_K}
\small{\hat{K}  \in   \underset{K}{\text{argmin}}  \left\{ \sum_{k,l = 1}^K \norm {A^{(k,l)}(\hat{Z}_K,K) -  
\Pi_{(1)} \lkr \Pi_{\hat{J}_K^{(k,l)}} \left(A^{(k,l)}(\hat{Z}_K, K)\right) \rkr}_{F}^2  + \Pen (n,\hat{J}_K,K)\right\}.}
\ee


\subsection{The Support of the Probability Matrix and the Penalty}
\label{sec:support}

Consider solution of optimization problem \eqref{eq:opt_ZK3} for a fixed value of $K$. 
If $\hat{Z}_K \in   \calM_{n,K}$ is a solution of \eqref{eq:opt_ZK},  then 
\small{\begin{align} 
  \hat{J}_K  & \in  \underset{J}{\text{argmin}}  
\left\{ \displaystyle \sum_{k,l = 1}^K \left\|A^{(k,l)}(\hat{Z}_K, K) -  \Pi_{(1)} \lkr \pijkl \lkr A^{(k,l)}(\hat{Z}_K, K)\rkr \rkr \right\|_{F}^2  
+ \Pen(n,J,K) \right\} \label{eq:opt_J_K} \\
 & \text{s.t.}\  A(\hat{Z}_K, K) = \scrPhZK^T A \scrPhZK, \    J^{(k,l)} = J_{k,l} \times J_{l,k}, 
\  J_{k,l} \equiv J_{k,l} (\hat{Z}_K, K). \nonumber 
\end{align} 
}
Observe that if the penalty term $\Pen(n,J,K)$ were not present in \eqref{eq:opt_J_K} or did not depend on a set $J$, then one would have 
$\hat{J}_K  = \brJ_K$  and $\hat{J}_K^{(k,l)} = \breve{J}_K^{(k,l)}$, where, by \eqref{eq:brJ},
$\brJ_K^{(k,l)}$ is the set of indices of nonzero rows and columns in $A^{(k,l)}(\hat{Z}_K, K)$. 
It is easy to see that    
\begin{align*}
& \pibrjkl \lkr A^{(k,l)}(\hat{Z}_K, K)\rkr = A^{(k,l)}(\hat{Z}_K, K),\\
& \Pi_{(1)} \lkr \pibrjkl \lkr A^{(k,l)}(\hat{Z}_K, K)\rkr \rkr = \Pi_{(1)} \lkr   A^{(k,l)}(\hat{Z}_K, K)  \rkr.
\end{align*} 
Hence, even if sparsity is not specifically enforced (as it happens in \cite {https://doi.org/10.1111/rssb.12410} where the penalty depends on $n$ and $K$ only),
one still obtains a sparse estimator $\hat{P}$  with the support  $\hat{J}_K  = \brJ_K$.

If the true number of clusters $K_*$ and the true clustering matrix $Z_* \in \calM_{n, K_*}$ were available, then 
the statement below shows that, with high probability, sets $J_* \equiv J_* (Z_*, K_*)$ and $\brJ (Z_*, K_*)$  would
coincide, provided   nonzero elements of matrix $P_*$ are above $C K_* \sqrt{\ln n/n}$ where $C$ is an absolute constant. 
Therefore, some zeros of the adjacency matrix correspond to the true zero probabilities of connections.

\begin{lem} \label{lem:brJ_Jstar}
Let $K_*^2 \leq n$ and the true matrix $P_*$ be such that $(P_*)_{i,j} =  0$ or $(P_*)_{i,j} > \varpi (n, K_*)$. 
If the community sizes are balanced, i.e., the sizes of the true communities are bounded below by $\tC_0 n/K_*$ for some $\tC_0 \in (0, 1]$, 
and 
\bes 
\varpi (n, K_*) \geq K_* \lkr \sqrt{\ln n} + \sqrt{t} \rkr \Big/ \lkr \tC_0 \sqrt{2 n} \rkr,
\ees  
then, with probability at least $1 - e^{-t}$, one has  $J_* (Z_*, K_*)= \brJ (Z_*, K_*)$.
\end{lem}

\begin{figure} [t]
\[\includegraphics[width = 15cm]{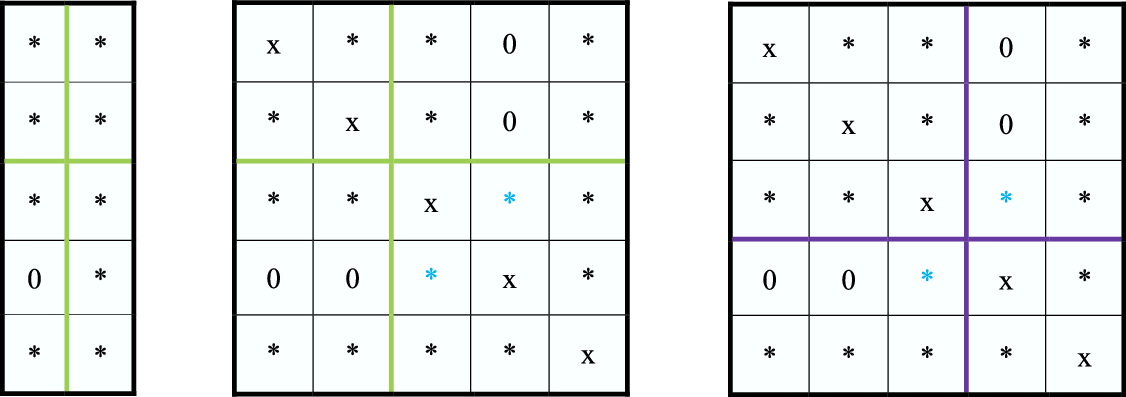}\]  
\caption{ {Zeros of the probability matrix with $n=5$ and $K_* =2$. 
Star symbols correspond to nonzero elements, the ``x'' symbols  stand for the diagonal elements that are unavailable, 
the thick lines correspond to clustering assignments.  
Left panel: matrix $\Lambda$ with $(J_*)_{1,1} = \{1,2 \}$, $(J_*)_{2,1} = \{3, 5\}$, 
$(J_*)_{1,2} = \{1,2\}$ and $(J_*)_{2,2} = \{3,4,5\}$.   
Middle panel: matrix $P_* (Z_*, K_*)$ with true clustering, $(\brJ_*)_{2,1}^c (Z_*) = \{4\}$,  
$\hat{P}_{i,j} (Z_*, K_*) = 0$ for $(i,j) \in   \{ (1,4), (2,4),   (4,1), (4,2)  \}$,
so that,   zero entries of the probability matrix are estimated by zeros. 
Right panel: matrix $P_* (\hat{Z}, K_*)$ with node~3 
erroneously placed into community~1.   The values of $(P_*)_{4,3}$ and $(P_*)_{3,4}$ are nonzero. 
If  $A_{3,4} = A_{4,3} =0$, then $\{4 \} \in \brJ_{2,1}^c (\hat{Z})$ and    $\hat{P}_{i,j} (\hat{Z}, K_*) = 0$
for    $(i,j) \in   \{ (1,4), (2,4), (3,4),   (4,1), (4,2), (4,3)  \}$, hence,  
zero entries of $P_*$ are still estimated by the identical zeros. 
However, if  $A_{4,3} = A_{3,4}  =1$, then zero elements $(P_*)_{4,1}$, $(P_*)_{4,2}$, 
$(P_*)_{1,4}$ and $(P_*)_{2,4}$ are estimated by positive values.}
}
\label{mn:fig0}
\end{figure}

Unfortunately, $K_*$ and $Z_*$ are unknown and, hence, $\hat{J}_K (Z,K)  = \brJ_K (Z,K)$ may not always be the best estimator.
In order  to understand this,  
consider, for example, the situation displayed in Figure~1 where $n=5$, $K_* =2$ and, under the true clustering, one has  $n_1=2$ and $n_2 =3$.
Vectors $\Lam_{2,1}$   has one zero element, so that $(J_*)_{1,1} = \{1,2 \}$, $(J_*)_{2,1} = \{3, 5\}$, 
$(J_*)_{1,2} = \{1,2\}$ and $(J_*)_{2,2} = \{3,4,5\}$ (left panel) leading to  
$(J_*)^{(1,1)} = \{(1,1), (1,2), (2,1), (2,2), \}$,
$(J_*)^{(2,1)} = \{ (3,1), (3,2), (5,1), (5,2) \}$, 
$(J_*)^{(1,2)} = \{ (1,3), (2,3), (1,5), (2,5) \}$ and   
$(J_*)^{(2,2)} = \{(3,3), (3,4), (3,5), (4,3),(4,4), (4,5), (5,3),\\ (5,4), (5.5)\}$ (middle panel). 
With the true clustering  (middle  panel), $(\brJ_*)_{2,1}^c (Z_*) = \{4\}$, so that 
$\hat{P}_{i,j} (Z_*, K_*) = 0$ for $(i,j) \in   \{ (1,4), (2,4),  (4,1), (4,2) \}$.
Hence,  zero entries of the probability matrix are estimated by zeros.

 Consider now the situation where the third node has been erroneously placed into community 1 by clustering matrix $\hat{Z}$ (right panel).  
Then, we   have $(J_*)^c_{2,1} = \{ 4 \}$ but $(\brJ_*)_{2,1}^c (\hat{Z})$  is an empty set.
If  $A_{3,4} = A_{4,3} =0$, then $\{4 \} \in \brJ_{2,1}^c (\hat{Z})$ and    $\hat{P}_{i,j} (\hat{Z}, K_*) = 0$
for    $(i,j) \in   \{ (1,4), (2,4),   (4,1), (4,2)  \}$, hence,  
zero entries of $P_*$ are still estimated by the identical zeros. 
However, if  $A_{4,3} = A_{3,4}  =1$, then it is possible that zero elements $(P_*)_{4,1}$, $(P_*)_{4,2}$, 
$(P_*)_{1,4}$ and $(P_*)_{2,4}$ are estimated by positive values. 
For example, if $A_{5,1}= 1$, $A_{5,2}= 1$ and $A_{5,3}= 1$, 
then $\hat{P}_{4,1} = 0.3536$ and $\hat{P}_{4,2} = 0.3536$ which leads to higher estimation errors than setting
$\hat{P}_{4,1} =  \hat{P}_{4,2} = 0$.
Therefore,  it is reasonable to introduce a penalty that will 
lead to trimming the support of $\hat{P} (Z,K)$.

One can consider two kinds of penalties here: separable and non-separable. 
We say that a penalty $Pen (n,J,K)$ is {\it separable} if for any $K$ and 
any clustering matrix $Z$ that partitions $n$ nodes into $K$ communities 
of sizes $n_k, k=1, \ldots, K$, one can write
\be \label{eq:sep_pen_def}
\Pen (n,J,K) = \Pen^{(0)} (n,J,K) +  \Pen^{(1)} (n,K) \  \mbox{with} \   
\Pen^{(0)} (n,J,K) = \sum_{l=1}^K \, \sum_{k=1}^K \mathscr{F} (|J_{k,l}|, n_k),
\ee
where $J_{k,l} \equiv J_{k,l} (Z,K)$. Otherwise, the penalty is {\it non-separable}. 
\\

\begin{lem} \label{lem:hatJ}
Let $(\hat{Z}_K , \hat{J}_K) $ be the solution of the optimization problem \eqref{eq:opt_ZK3}.
If   $Pen (n,J,K)$ is an increasing function of $|J|$ (for a non-separable penalty) or of $|J_{k,l}|, k,l = 1,\ldots, K$ 
(for a  separable penalty), then   
\be \label{eq:hatJbrJ}
\hat{J}_{k,l} (\hat{Z}_K , K) \subseteq \brJ_{k,l} (\hat{Z}_K , K) \subseteq (\brJ_*)_{k,l} (\hat{Z}_K , K), \quad
\hat{J} (\hat{Z}_K , K) \subseteq \brJ (\hat{Z}_K , K) \subseteq \brJ_* (\hat{Z}_K , K). 
\ee
\end{lem}



\section{The Errors of  Estimation and Clustering}
\label{sec:errors}

\subsection{The penalty}
\label{sec:penalty} 

 In what follows, we consider the separable and the non-separable penalties   of the form \eqref{eq:sep_pen_def} with
the common  $\Pen^{(1)} (n,K)$ term, i.e.
\be \label{eq:penalties}
\Pen^{(a)} (n,J,K) = \Pen^{(0,a)} (n,J,K) + \Pen^{(1)} (n,K), 
\ee
 where a =s for the separable penalty and a = ns for the non-separable one, and
\beqn 
\label{eq:pen0s}
\Pen^{(0,s)} (n,J,K) & = & \beta_1 \di\sum_{k,l = 1}^K  |J_{k,l}|  \ln (n_k e/|J_{k,l}|) + \beta_2 K \di\sum_{k  = 1}^K  \ln n_k \\
\label{eq:pen0ns}
\Pen^{(0,ns)} (n,J,K) & = & \beta_1   |J |  \ln (n K e/|J|) + 2 \beta_2  \ln n \\
\label{eq:pen1}
\Pen^{(1)} (n,K) & = & \beta_2   [n \ln K +   \ln n].
\eeqn 
Here, the separable penalty corresponds to  
$\mathscr{F} (|J_{k,l}|, n_k) = \beta_1 |J_{k,l}|\,  \ln (n_k e/|J_{k,l}|) + \beta_2  \ln n_k $
and the exact expressions for $\beta_1$ and $\beta_2$ are given in the proof of Theorem~\ref{th:oracle}.

In the next two sections, we shall provide    upper bounds for the   errors 
of the solution  of optimization problem \eqref{eq:opt_main} with the separable or the non-separable penalty \eqref{eq:penalties}, 
as well as upper bounds for the clustering error in the case of the separable penalty. 
While   the separable penalty has some valuable properties (see Lemma~\ref{lem:hatJ}), 
the non-separable penalty is much easier to interpret. Fortunately,
as the statement below shows, under very nonrestrictive  conditions, 
the penalties are within a constant factor of each other.

\begin{lem} \label{lem:penalty}
If $n \geq 8$ and $K \leq \sqrt{n/\ln n}$, then
\be \label{eq:pen_rel}
\Pen^{(ns)} (n,J,K) < (2 + \beta_1/\beta_2)\, \Pen^{(s)} (n,J,K) <  2\, (2 + \beta_1/\beta_2)\, \Pen^{(ns)} (n,J,K).
\ee 
\end{lem}


\subsection{The Estimation Errors}
\label{sec:est_err}

\begin{thm} \label{th:oracle}
Let $(\hat{\Theta}, \hat{Z},\hat{J},\hat{K})$ be a solution of optimization problem  \eqref{eq:opt_main}
with the  penalty  defined  in  \eqref{eq:penalties}.
Construct the estimator $\hat{P}$  of $P_*$ of the form 
 \be \label{eq:P_total_est}      
\hat{P}  =\mathscr{P}_{\hat{Z},\hat{K}}\ \hat\Theta(\hat{Z},\hat{J},\hat{K})\ \mathscr{P}_{\hat{Z},\hat{K}}^T 
\ee
where $\mathscr{P}_{\hat{Z},\hat{K}}$ is the permutation matrix corresponding to  $(\hat{Z},\hat{K})$. 
%
Then, for any $t >0$ and some absolute positive constants $\gamma$ and $\tilde{C}$, one has 
\be \label{eq:oracle}
\PP \lfi  n^{-2}\, \|\hat{P}  -P_{*}\|_F^2  \leq   
n^{-2}\, H_0 \, \Pen (n,J_{*},K_{*})     + n^{-2}\, \tilde{C} t  \rfi \geq  1 - 3 e^{-t},
\ee 
\be \label{eq:oracle Expectation}
n^{-2}\,   \EE \|\hat{P}  - P_{*}\|_F^2  \leq   n^{-2}\, H_0 \, \Pen (n,J_{*},K_{*}) +  3 n^{-2}\, \tilde{C}.  
\ee 
The exact expressions for $H_0$ and $\tilde{C}$ are given in the proof of Theorem~\ref{th:oracle}. 
\end{thm}

\medskip
\medskip

\noindent
Observe that, due to Lemma~\ref{lem:penalty}, the separable and non-separable penalties are within a constant factor of each other, so that
 Theorem~\ref{th:oracle} implies that  
the estimation error   is proportional to $\Pen (n,J_{*},K_{*})$ where
\be \label{eq:estim_er}
\Pen (n,J,K) \asymp  \Pen^{(ns)} (n,J,K) \asymp 
  n \ln K +  |J|\, \ln (n K e/|J|)   +  \ln n.
\ee
The first term in \eqref{eq:estim_er} is due to the clustering errors, the second term quantifies the difficulty 
of finding  $|J|$ nonzero elements among $nK$ elements of matrix $\Lambda \in [0,1]^{n \times K}$ and estimating them, while 
the  term $\ln n \asymp \ln (nK)$ stands for the difficulty of finding the cardinality of the set $|J|$, and it is always   
dominated by the first two terms in  \eqref{eq:estim_er}.

Since each node  is connected to at least one community   with a nonzero probability, one has $n \leq |J| \leq n K$.
In the (non-sparse) PABM, $|J| = nK$ and the second term in \eqref{eq:estim_er} is always asymptotically larger
than the other two terms, as $n \to \infty$. In SPABM, the second term in \eqref{eq:estim_er} dominates the first term only if $K=1$ 
or $|J|/n \to \infty$ as $n \to \infty$. However, if $K>1$ and $|J| \asymp n$, then both terms are of the equal asymptotic order. 
If  $K \to \infty$  and $|J| \asymp n$ as $n \to \infty$, then 
SPABM has the error $O( n \ln K)$ which is asymptotically smaller than  $O(n K)$ error of PABM.


\subsection{Detectability of clusters}
\label{sec:clust_detect} 

In order one can detect clusters, the vectors $\Lam^{(k,l)}$, $l=1, \ldots, K$,
should be sufficiently different for every  $k=1, \ldots, K$. Assume that $K=K_*$ is known and that 
the following condition holds.
\\

\noindent
{\bf Assumption A1. \ }  For any $k=1,\ldots, K$, vectors ${\Lambda}^{(k,1)}, \ldots,  {\Lambda}^{(k,K)}$ 
are linearly independent.
\\

\noindent
Under Assumption A1, the true clusters are detectable.

\begin{lem} \label{lem:detect}
Let $Z_* \in \calM_{n, K}$ be  the true clustering matrix,  and $Z \in \calM_{n, K}$
 be an arbitrary clustering matrix. 
Let $J_* = J_*(Z_*)$ be  the true set of indices of nonzero elements, and $\brJ_* = \brJ_* (Z)$  be  
the set of indices of nonzero elements, defined in \eqref{eq:brJ}, which is  associated with a clustering matrix $Z \in \calM_{n, K}$.
If Assumption {\bf A1}    holds and the network is connected, then
\be \label{eq:detect}
\small{ \sum_{k,l = 1}^K\, \norm{P_*^{(k,l)}(Z_*) -  \Pi_{(1)} \lkr \Pi_{J_*^{(k,l)}} (P_*^{(k,l)}(Z_*) \rkr }_{F}^2 
\leq  \sum_{k,l = 1}^K \, \norm{P_*^{(k,l)}(Z) -  \Pi_{(1)} \lkr \Pi_{\brJ_*^{(k,l)}}(P_*^{(k,l)}(Z) \rkr }_{F}^2}
\ee 
where, for any matrix $B$, $\Pi_{(1)} (B)$ is its rank one approximation and $\Pi_J$ is its projection on the set of indices defined by $J$. 
Moreover, equality in \eqref{eq:detect} occurs if and only if matrices  $Z$ and $Z_*$ coincide up to a permutation of columns.
\end{lem}


\subsection{The Clustering Errors}
\label{sec:clust_err}

In order to evaluate the clustering error when clustering is applied to the adjacency matrix, 
we assume that the true number of classes $K = K_*$ is known.
Then $\hat{Z}  \equiv \hat{Z}_K$ is a solution of the optimization problem \eqref{eq:opt_ZK3}.

Let $Z_* \in \calM_{n, K}$ be the true clustering matrix and  $Z_* \in \calM_{n, K}$ 
be any other clustering matrix.  Then the proportion of misclustered nodes can be evaluated as 
 \be\label{eq:misclustered}
 \Err(Z, Z_{*}) =  (2n)^{-1}\, \underset{\mathscr{P}_K \in \mathcal{P}_K} {\min}  \|Z \mathscr{P}_K - Z_{*}\|_1
= (2n)^{-1}\,  \underset{\mathscr{P}_K \in \mathcal{P}_K} {\min}  \|Z \mathscr{P}_K - Z_{*}\|_F^2  
 \ee
where $\mathcal{P}_K$ is the set of permutation matrices  $\mathscr{P}_K: \{ 1,2,\cdots,K\} \longrightarrow \{ 1,2,\cdots,K\}$.
Let 
\be\label{eq:def;Lambda}
 \Upsilon(Z_{*},\del_n) = \lfi Z \in \mathcal{M}_{n,K } :  (2n)^{-1}\,  \underset{\mathscr{P}_K \in \mathcal{P}_K} 
{\min}  \|Z \mathscr{P}_K - Z_{*}\|_1 \geq \del_n \rfi 
 \ee
be the set of clustering matrices with the proportion of misclassified  nodes being at least  $\del_n \in (0,1)$.

The success of clustering in \eqref{eq:opt_ZK3} relies upon the fact that matrix $P_*$ is a collection of $K^2$ rank one blocks,
so that the operator and the Frobenius norms of each block are the same.  On the other hand, if clustering were incorrect,
the ranks of the blocks would increase which would lead to the discrepancy between their operator and Frobenius norms. 
In particular, the following statement is true.


\begin{thm}  \label{th:clust}
Let  $K = K_*  \geq 2$ be the true number of clusters. $Z_* \in \calM_{n, K}$ be the true clustering matrix 
and Assumption {\bf A1}  hold.
Let $J_* = J_*(Z_*)$ be  the true set of indices of nonzero elements, and $\brJ_* = \brJ_* (Z)$  be  
the set of indices of nonzero elements, defined in \eqref{eq:brJ}, which is  associated with a clustering matrix $Z \in \calM_{n, K}$.
Let $\hat{Z}  \equiv \hat{Z}_K$ be a solution of the optimization problem \eqref{eq:opt_ZK3} and $\del_n \to 0$ as $n \to \infty$.
If there exists  $\alpha_n  \in (0,1/2)$   and absolute positive constants
 $H_1$ and $H_2$, independent of $K$, $n$,  $J_*$, $\brJ_* (Z)$,  $\del_n$ and $\alpha_n$,
 such that 
 \begin{align} 
\| P_{*}\| _{F}^2 & \geq    \underset{Z \in \Upsilon(Z_{*},\del_n)} {\max} 
\lkv (1 + \alpha_n)\, \sum_{k,l = 1}^K \|  P_{*}^{(k,l)} (Z) \|_{op}^2  + 
  \frac{H_1}{\alpha_n}\,  |\brJ_* (Z)| \ln \lkr \frac{nKe}{|\brJ_* (Z)|} \rkr \rkv \nonumber\\
& + \frac{H_1}{\alpha_n}\,  \lkr |J_*| + n \ln K  \rkr + H_2 |J_*| \ln \lkr \frac{nKe}{|J_*|} \rkr  \label{eq:cond_check}
\end{align}
then, with probability at least $1 - 2 K^{-n}$, the proportion of the nodes, misclassified  by $\hat{Z}$,
is at most $\del_n$.
\end{thm}


\begin{ex}\label{ex1}
{\rm
In order to see what condition \eqref{eq:cond_check} means, we consider a simple example.
We study the sparse PABM    with $K=2$,  
and $Z_* \in \calM_{n,2}$ with equal size communities $N=n/2$.
Assume that $\Lam^{(k,k)} = \sqrt{a}\, 1_N$, $k=1,2$, while   elements $\Lam^{(k,l)}_i$ of vectors  $\Lam^{(k,l)}$, $k \neq l$, are equal to
$\sqrt{b}$ if  $i \in J_k$, $k=1,2$,  and   equal  to zero otherwise. 
Examine the case of an assortative network, where $a \equiv a_n$, $b \equiv b_n$ and  $b/a =  \rho \equiv  \rho_n \leq 1$.
Denote   $J = J_1 \cup J_2$ and note  that the cardinality 
of the set of nonzero elements of matrix $\Lam$ is equal to $2N + |J|$ with $|J| = |J_1|+ |J_2|$.
Denote the overall proportion of nonzero entries in vectors $\Lam^{(1,2)}$ and $\Lam^{(2,1)}$ by $\gam$, 
and the proportion of zero entries in vectors $\Lam^{(1,2)}$ and $\Lam^{(2,1)}$ by $s$:
\bes 
\gam   = |J|/n = (|J_1|+ |J_2|)/(2N), \quad
s = 1- \gam.  
\ees
Below we examine what condition \eqref{eq:cond_check} of Theorem~\ref{th:clust}  means for different values 
of $s$ and $\rho$.  Assume that the connection probabilities are not too small, specifically, 
that 
\be \label{eq:an_cond}
\lim_{n \to \infty} n a_n^2 = \infty.
\ee

Let $\del_n \equiv \del =  \del_1 + \del_2$.
Let $Z  \in \Upsilon(Z_{*},\del_n) \subset \calM_{n,2}$ be an arbitrary incorrect clustering matrix and, according to $Z$, 
$\cNk = N \del_k$ nodes are moved erroneously from class $k$ to class $l$,
$l \neq k$, and  $\tNk = N (1-\del_k)$ nodes remain  correctly in class $k$.
Then, according to $Z$, community $k$ has $\tNk + \cNl$ nodes, $k=1,2$, $k \neq l$, and the proportion  of misclassified nodes
is equal to $(\del_1 N + \del_2 N)/n = \del/2$. 
Denote  the subsets of nodes corresponding to nonzero elements of vector $\Lam^{(k,l)}$,  that correctly stay 
in class $k$ and those that are misclassified into community $l$, $l \neq k$, by  $\tJk$ and $\cJk$, respectively. 
Then $J_k = \tJk \cup \cJk$, $k=1,2$.
Denote
\be \label{eq:tcbek}
\tbek = |\tJk|/|\tNk|, \ \ \cbek = |\cJk|/|\cNk|,\quad k=1,2,
\ee 
and note that $\tbek, \cbek \in [0,1]$. 
Then,  for any $Z \in \calM_{n,2}$ with equal class sizes and the proportion of misclassified nodes being $\del/2$, one has
\be \label{eq:cond_example}
 \|P_{*}\|_{F}^2 -  (1 + \alpha_n)\,  \sum_{k,l = 1}^2 \|P_{*}^{(k,l)} (Z)\|_{op}^2  
 \geq \cC\, a^2 n^2 (\Del_n (Z) -  16\, \alpha_n), 
\ee
where $\cC$ is an   absolute constant, $\alpha \equiv \alpha_n$  and
\be   \label{eq:Deln}
\Del_n (Z) = \del_1^2\, (1 - \rho_n^2\, \cbe_1^2\, \tbe_2^2)^2 + \del_2^2\, (1 - \rho_n^2\, \tbe_1^2\, \cbe_2^2)^2.
\ee  
The proof of the inequality \eqref{eq:cond_example} is given in the Appendix. 

Note that, in this example,  the right hand side of \eqref{eq:cond_check} reduces to $H_1\, n\, \alpha_n^{-1} + H_2 n$,
so   we need to show that
\be \label{eq:cond_check1}
a^2 n^2  \Del_n (Z) \geq  16\, a^2 n^2 \, \alpha_n  
 +   \tH_1\, n\, \alpha_n^{-1} + \tH_2 n,
\ee 
 for some $\alpha_n  \in (0,1/2)$   and absolute positive constants
 $\tH_1$ and $\tH_2$.  It is easy to see that the right hand side of \eqref{eq:cond_check1}
is minimized by $\alpha_n = C_{\alpha} \, /(a_n \sqrt{n}) \in (0,1/2)$, and \eqref{eq:cond_check1} appears as
\be \label{eq:cond_check2}
 a_n  \sqrt{n}\   \Del_n (Z) \geq \tH_3  
\ee
for some absolute positive constant $\tH_3$.  Below, we examine when this condition can be satisfied for  $\del_n \to 0$ as $n \to \infty$.

First, we consider the case when $s=0$, so that 
$\ga=1$ and  there is no structural sparsity.  In this case, $\cbek = \tbek =1$, $k=1,2$,
and, due to  $\del_1^2 + \del_2^2 \geq (\del_1 + \del_2)^2/2 = \del^2/8$, one obtains from \eqref{eq:Deln} that  
$\Del_n (Z) \asymp \del_n^2  (1 - \rho_n^2)^2$.  Hence, \eqref{eq:cond_check2} becomes 
$ a_n  \sqrt{n}\, \del_n^2  (1 - \rho_n^2)^2 \geq \tH_3$, so that 
\be \label{eq:cler1}
\del_n^2 \asymp \lkv n a_n^2 (1 - \rho_n^2)^2 \rkv^{-1} \to 0 \quad \mbox{if} \quad
n a_n^2 (1 - \rho_n^2)^2 \to \infty \quad (n \to \infty).  
\ee 
The latter implies that either $a_n$ should be asymptotically larger than $n^{-1/2}$
or the ratio $\rho_n = b_n/a_n$ should be separated from one.

Now, consider $s>0$, so that $\ga < 1$. In this case we need the minimal possible value of $\Del_n(Z)$ 
over $Z \in \Upsilon(Z_{*},\del_n)$ to satisfy condition \eqref{eq:cond_check2}.
To formalize this notion, we introduce
\begin{align} 
  \widehat{F} (\ga, \del, \rho, a, n)  & = \min \lfi \del_1^2\, (1 - \rho_n^2\, \cbe_1^2\, \tbe_2^2)^2 
                                              + \del_2^2\, (1 - \rho_n^2\, \tbe_1^2\, \cbe_2^2)^2 \rfi \nonumber \\
   \mbox{s.t.} \quad & 0 \leq \tbe_k \leq 1,\ 0 \leq \cbe_k \leq 1,\ \tbek, \cbek \ \mbox{given by}\ \eqref{eq:tcbek}, \ \ k=1,2,\nonumber \\  
  &  \del_k \geq 0,\ k=1,2,\ \del_1 + \del_2 = \del \leq 1/2 \nonumber \\
  & \tbe_1 (1 - \del_1) + \tbe_2 (1 - \del_2) + \cbe_1 \del_1 + \cbe_2 \del_2 = 2 \ga \label{eq:beta_cond}
\end{align}
In order the proportion of clustering errors is bounded above by $\del_n \to 0$, one needs 
\be \label{eq:fin_dcond}
\widehat{F} (\ga, \del_n, \rho_n, a_n, n) \asymp \lkr a_n^2 n\rkr^{-1/2}, \quad (n \to \infty).
\ee

Consider the case when $s <1/2$, so that $1/2  < \ga < 1$. 
If $\del_n \to 0$, then, for $n$ large enough, one has $\del_n \leq 2(\ga - 1/2)$ and, hence, $2 \ga \geq 1 + \del_n$.
Set $\del_1=\del$, $\del_2 = 0$, $\cbe_1 = \tbe_2 =1$,
$\tbe_1 = [2 \ga - (1 + \del)]/(1-\del)$, $\cbe_2 = 0$. 
It is easy to verify that  conditions in \eqref{eq:beta_cond} hold, so that 
$\widehat{F} (\ga, \del, \rho, a, n) \leq \del^2  (1 - \rho_n^2)^2$ and 
condition \eqref{eq:fin_dcond} is equivalent to \eqref{eq:cler1}, that occurs when there is no structural sparsity ($s=0$).

Now, let $s > 1/2$, so that $0 < \ga < 1/2$. 
Let $d = (1 - 2\ga)/2$  and, if  $\del_n \to 0$, then, for $n$ large enough, one has $\del_n \leq d$.
Let $\ga_0 = (1 - 2d)/(1 -d)<1.$ Then, $2 \ga/(1 - \del_n)  \leq \ga_0$. By \eqref{eq:beta_cond},
obtain $\tbek (1 - \del_k) \leq 2\ga$ and, hence, $\tbek \leq 2\ga/(1 - \del_k) \leq \ga_0$, $k=1,2$.
Consequently, due to $\tbek, \cbek \leq 1$, obtain 
\bes
\widehat{F} (\ga, \del_n, \rho_n, a_n, n) \geq (\del_1^2 + \del_2^2) (1 - \rho_n^2 \ga_0^2)^2 \geq \del_n^2 (1 - \ga_0^2)^2 /2.
\ees
Since $\ga_0$ is a non-asymptotic quantity,   condition \eqref{eq:fin_dcond} holds  for some $\del_n \to 0$ 
as $n \to \infty$, whenever assumption \eqref{eq:an_cond} is satisfied. 
Therefore, if $s>1/2$, one has $\del_n \to 0$ even if $\rho_n \to 1$ as $n \to \infty$.

The sparsity proportion of $s = 1/2$ constitutes the so called ``elbow'' value, so the difficulty of clustering 
varies significantly for $s<1/2$ and $s>1/2$.   Analysis of the conditions that ensure $\del_n \to 0$ when $s=1/2$ 
requires more sophisticated tools,   so we do not study $s=1/2$  in this paper. 
} 
\end{ex}

\begin{rem}\label{rem:const_ab}
{\bf Non-constant connection probabilities. }{\rm
We remark that consideration of constant values for elements of vectors $\Lam^{(k,k)}$ and $\Lam^{(k,l)}$, $k,l=1,2$, 
$k \neq l$, is motivated by showing a clear pattern of the impact of sparsity on the clustering precision. 
Assumption that in-cluster and out-of-cluster connection probabilities  take constant values  are quite common in stochastic networks
literature (see, e.g.,   \cite{JMLR:v18:16-480},  \cite{doi:10.1137/19M1257135}, \cite{10.1214/19-AOS1854} and 
\cite{ndaoud2020improved} among others). Indeed, if $\Lam^{(k,k)}_i \geq \sqrt{a_n}$,  and $\Lam^{(k,l)}_i \leq \sqrt{b_n}$ 
if  $i \in J_k$, $k=1,2$, $k \neq l$, and   equal  to zero otherwise, where  $b_n/a_n =     \rho_n \leq 1$,
then conclusion of Example~\ref{ex1} that $\del_n \to 0$ as $n \to \infty$ is still true,  provided condition \eqref{eq:an_cond} holds
and $s >1/2$. However, in the case of $s < 1/2$, $\del_n$ may  tend to zero even if $s<1/2$, depending
on the exact values of components of vectors $\Lam^{(k,k)}$ and $\Lam^{(k,l)}$. 
Studying the case of constant probabilities allowed us to show the benefits of structural sparsity 
more clearly.
}
\end{rem}


\section { Implementation of Clustering }
\label{sec:clustering}

In Section~\ref{sec:est_clust}, we obtained an estimator $\hat{Z}$ of the true clustering matrix $Z_*$ 
as a solution of optimization problem \eqref{eq:opt_ZK}. Minimization in \eqref{eq:opt_ZK} is somewhat similar to modularity maximization
in \cite{Bickel21068} or \cite{zhao2012consistency} in the sense that 
modularity maximization as well as minimization in \eqref{eq:opt_ZK} 
are NP-hard, and, hence,  require some relaxation in order to obtain  an implementable clustering solution.

In the case of the SBM and the DCBM,  possible relaxations  include semidefinite programming 
(see, e.g., \cite{DBLP:journals/corr/AminiL14} and references therein),
variational methods (\cite{celisse2012}) and  spectral clustering and its versions 
(see, e.g.,    \cite{joseph2016}, \cite{lei2015}  and   \cite{rohe2011spectral} among others).
Since in the case of SPABM, columns of matrix $P_*$ that correspond to nodes in the same class are neither identical, nor proportional,
direct application of spectral clustering to matrix $P_*$  does not deliver the partition of the nodes.
However, it is easy to see that the columns of   matrix $P_*$ that correspond to nodes in the same community, form a matrix with $K$ rank-one blocks,
hence, those columns lie in the subspace of the dimension  at most $K$. Therefore, matrix $P_*$ is constructed of $K$ clusters of 
columns (rows) that lie in the union of $K$ distinct subspaces, each of the dimension $K$.
For this reason, the subspace clustering presents a technique for 
obtaining a fast and reliable solution of optimization problem~\eqref{eq:opt_ZK} (or \eqref{eq:opt_ZK3}).

Subspace clustering has been widely used in computer vision and, for this reason, it is a very
well studied and developed technique.
Subspace clustering is designed for separation of points that lie in the union of subspaces. 
Let $\{ X_{j} \in \R^{D} \}_{j=1}^{n} $ be a given set of points drawn from an unknown union of 
$K \geqslant 1$ linear or affine subspaces $\{ S_{i} \}_{i=1}^{K} $ of unknown dimensions 
$d_{i}= \text{dim}(S_{i})$, $0<d_{i} <D$, $i=1,...,K$. In the case of linear subspaces, the subspaces can be described as 
$S_{i}=\{ \bm{x} \in \R^{D} : \bm{x}=  \bm{U}_{i}\bm{y} \}$, $i=1,...,K,$ 
where $\bm{U}_{i} \in \R^{D \times d_{i}}$ is a basis for subspace $S_{i}$ 
and $\bm{y} \in \R^{d_{i}}$ is a low-dimensional representation for point $\bm{x}$. 
The goal of subspace clustering is to find the number of subspaces $K$, their dimensions 
$\{ d_{i} \}_{i=1}^{K}$, the subspace bases $\{ \bm{U}_{i} \}_{i=1}^{K}$, and the segmentation 
of the points according to the subspaces.

Several methods have been developed to implement subspace clustering such as  algebraic methods 
(\cite{inproceedings}, \cite{Ma:2008:ESA:1405158.1405160}, \cite{vidal2005generalized}), 
iterative methods (\cite{P.AgarwalandN.Mustafa2004}, \cite{Bradley:2000:KC:596077.596262},  \cite{tseng2000nearest}), 
and spectral clustering based methods (\cite{Vidal:2009aa},  \cite{Elhamifar:2013:SSC:2554063.2554078}, 
\cite{Favaro:2011:CFS:2191740.2191857},  \cite{Liu:2013:RRS:2412386.2412936}, 
\cite{Liu2010RobustSS},  \cite{soltanolkotabi2014},  
 \cite{vidal2011subspace}). 
In this paper, we   use the latter group of techniques.

Spectral clustering algorithms rely on  construction of  an affinity matrix 
whose entries are based on some distance measures between the points. 
In particular, in the case of the SBM, adjacency matrix itself serves as the affinity matrix, while for the DCBM, 
the affinity matrix is obtained by normalizing rows/columns of $A$. 
In the case of the  subspace clustering problem, one cannot use the typical distance-based affinity
because two points could be very close to each other, but lie in different subspaces, while they 
could be far from each other, but lie in the same subspace. One of the solutions is to construct 
the affinity matrix using self-representation of the points with the expectation that a point is more likely to 
be presented as a linear combination of points in its own subspace rather than from a different one.  
A number of approaches such as Low Rank Representation  (see, e.g., \cite{Liu:2013:RRS:2412386.2412936},
\cite{Liu2010RobustSS}) and Sparse Subspace Clustering  (see, e.g., 
\cite{Elhamifar:2013:SSC:2554063.2554078},  \cite{Vidal:2009aa})
have been proposed in the past decade   for the solution of this problem.  

In this paper, we use Sparse Subspace Clustering (SSC) since it allows one to take advantage of the knowledge that, for a given $K$,
columns of matrix $P_*$  lie in the union of $K$ distinct subspaces, each of the dimension at most $K$. 
If matrix $P_*$ were known, the weight matrix $W$ would be based on writing every data point 
as a sparse linear combination of all other points by solving the following optimization problem
\begin{equation}  \label{mn:opt_prob2}
\mathop\text{min}_{W_{j}}  \|W_{j}\|_{1}  \hspace{5mm}   \mbox{s.t.}  \hspace{5mm}  (P_*)_{j}=\sum_{k \ne j} W_{kj} (P_*)_{k}
\end{equation}
In the case of data contaminated by noise, the SSC algorithm does not attempt to write  
data as an exact linear combination of other points. 
Instead, SSC can be built upon   the solution of   the elastic net problem 
\begin{equation} \label{mn:Elastic_Net}
\widehat{W}_j \in \underset{W_{j}}{\text{argmin}}  
\lfi   \lkv \frac{1}{2} \|{A_{j}-AW_{j}}\|_{2}^{2} + \gamma_1 \|W_{j}\|_{1} + \gamma_2 \|W_{j}\|_{2}^{2}\rkv 
\quad   \mbox{s.t.}  \quad W_{jj}=0  \rfi, \quad j=1,...,n,
\end{equation}
where $\gamma_1, \gamma_2 > 0$ are tuning parameters. The quadratic term stabilizes the LASSO problem by making the problem strongly convex, 
and therefore it has a unique minimum.

We solve  \eqref{mn:Elastic_Net} using   the LARS algorithm
\cite{Efron04leastangle} implemented  in  SPAMS Matlab toolbox (see \cite{mairal2014spams}).  
Given $\widehat{W}$,  the affinity matrix is defined as $|\widehat{W}| + |\widehat{W}^{T}|$ where, for any matrix $B$, matrix $|B|$
has absolute values of elements of $B$ as its entries.
The class assignment (clustering matrix) $Z$ is then obtained by applying spectral clustering to $|\widehat{W}| + |\widehat{W}^{T}|$. 
We elaborate on the implementation of the SSC in Section \ref{sec:simulations}.


\section {Simulations and Real Data Examples}
\label{sec:sim_real}

\subsection{Simulations on Synthetic Networks}
\label{sec:simulations}

\begin{figure}
\centerline{ \includegraphics[width = 17cm]{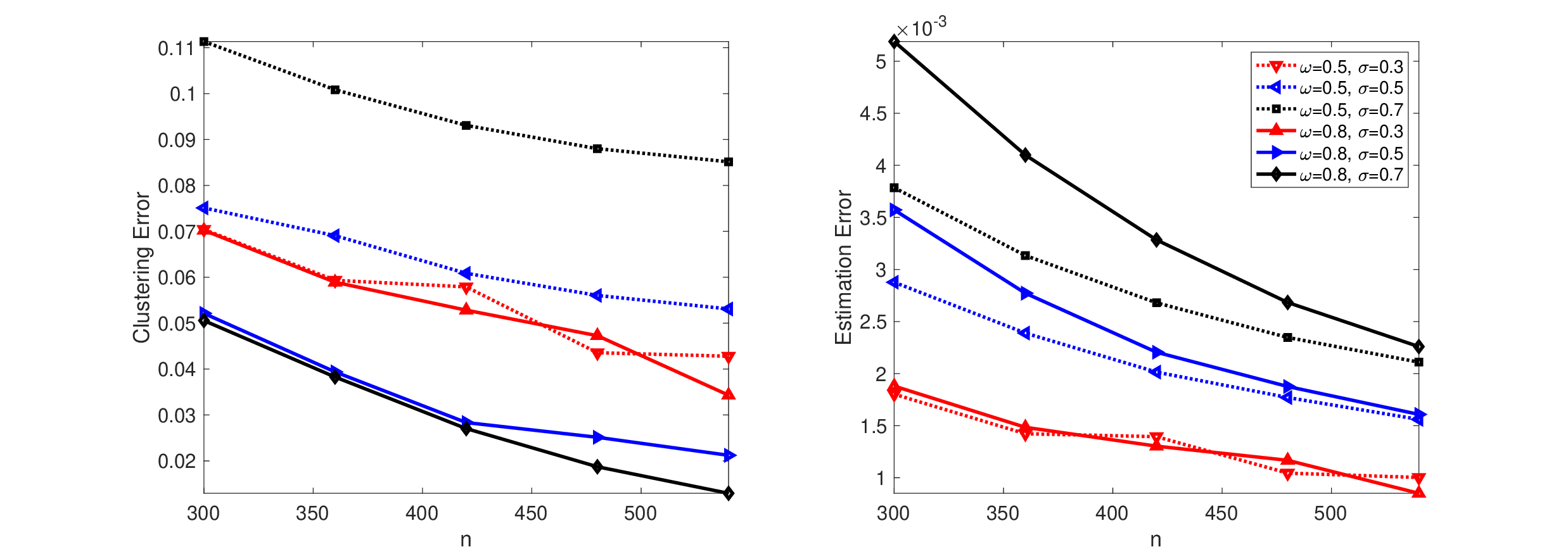}  }
\centerline{ \includegraphics[width = 17cm]{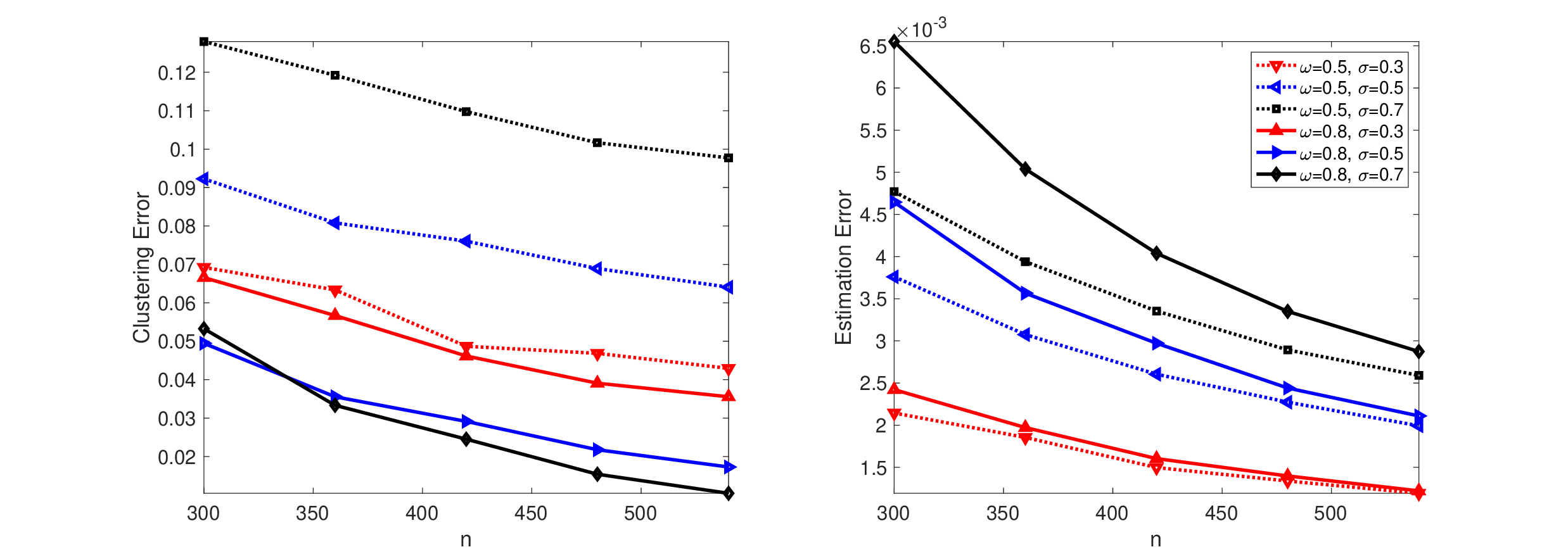}  }
\centerline{ \includegraphics[width = 17cm]{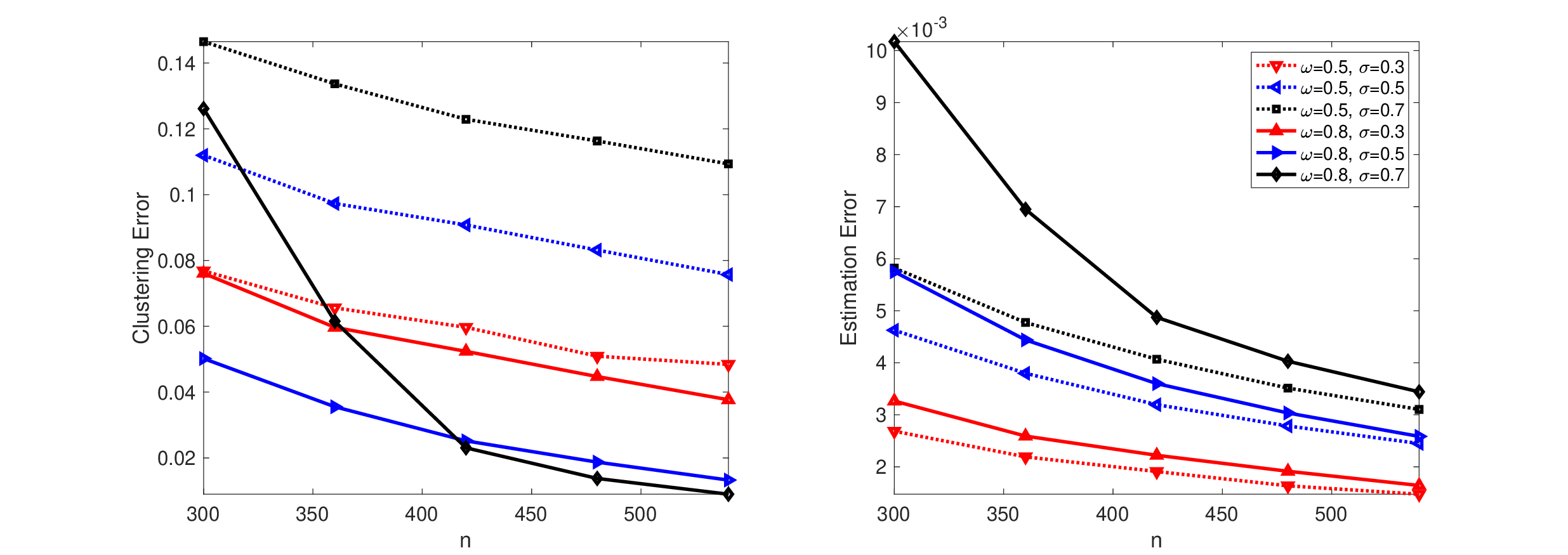}  }

\caption{ The clustering errors $\Err (\hat{Z},Z)$ defined in \eqref{eq:misclustered}  (left panels)
and the estimation errors $n^{-2}\, \|{\hat{P}-P}\|_{F}^{2}$ (right panels) for $K=4$ (top), $K=5$ (middle) and 
$K=6$ (bottom) clusters. The errors are evaluated over 100 simulation runs. 
The number of nodes ranges from $n=300$ to $n=540$ with the increments of 60. 
Dashed lines represent the results for $\om=0.5$ and solid lines represent the results for $\om=0.8$;
$\sigma=0.3$  (red), $\sigma=0.5$ (blue) and $\sigma=0.7$ (black).
}
\label{mn:fig1}
\end{figure}

In this section we evaluate the performance of our method using synthetic networks. 
We assume that the number of communities (clusters) $K$ is known and for simplicity consider 
a perfectly balanced model with $n/K$ nodes in each cluster.
We generate each network  from a random graph model with a  symmetric probability  matrix $P$
given by the SPABM model with a clustering matrix $Z$ and a block matrix $\Lambda$.

To generate synthetic networks, we start by producing a block matrix $\Lambda$ in \eqref{eq:Lambda} 
with random entries between 0 and 1. We use a parameter $\sigma$ as the proportion of nonzero entries in matrix $\Lambda$ to control the sparsity of networks. To do that, we set $\left \lfloor{nK\sigma}\right \rfloor$ smallest non-diagonal entries of $\Lambda$ zero.  
Then we multiply the non-diagonal blocks of $\Lambda$ by $\omega$, $0<\omega <1$,  to ensure that most nodes in the same community have 
larger probability of interactions. As a result, matrix $P(Z,K)$ with blocks $P^{(k,l)}(Z,K) =\Lambda^{(k,l)}  (\Lambda^{(l,k)})^{T}$,
$k,l = 1, \ldots, K$, has larger entries mostly in the diagonal blocks than in the non-diagonal blocks and some zero rows (columns) in the non-diagonal blocks. 
The parameter $\om$ is the heterogeneity parameter. Indeed, if $\om=0$, 
the matrix $P_*$ is strictly block-diagonal, while in the case of $\om =1$, there is no difference between entries in diagonal and nonzero entries in non-diagonal blocks.
Next, we generate a random clustering matrix $Z \in \calM_{n,K}$ corresponding to the case of equal community sizes and the 
permutation matrix $\scrPZK$ corresponding to the clustering matrix $Z$. Subsequently, we scramble rows and columns 
of $P(Z, K)$ to create the probability matrix  $P=\scrPZK P(Z,K) \scrPZK^T$.  
Finally we generate the lower half of the adjacency matrix $A$ as independent Bernoulli variables  
$A_{ij} \sim \text{Ber}(P_{ij})$, $i=1, \ldots, n, j=1, \ldots, i-1$, and set  $A_{ij} = A_{ji}$ when $j >i$. 
In practice, the diagonal elements of matrix $A$ are unavailable, so we estimate $\diag (P)$ without their knowledge.

\begin{figure}
\centerline{ \includegraphics[width = 17cm]{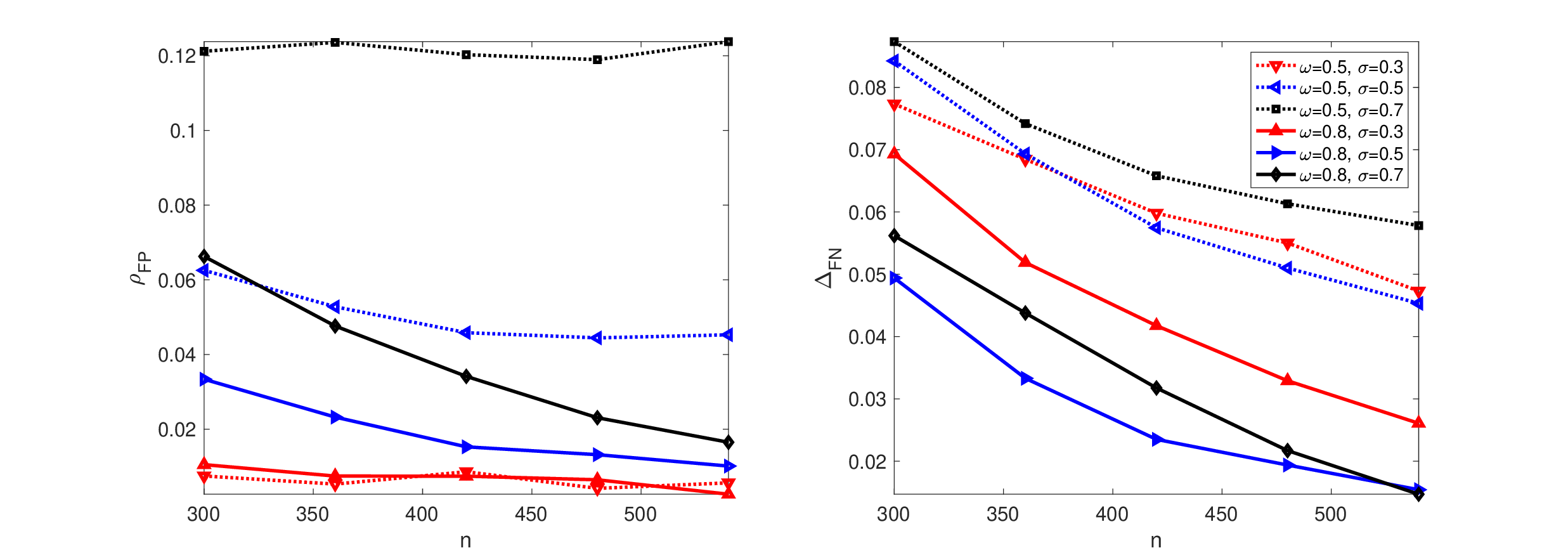}  }
\centerline{ \includegraphics[width = 17cm]{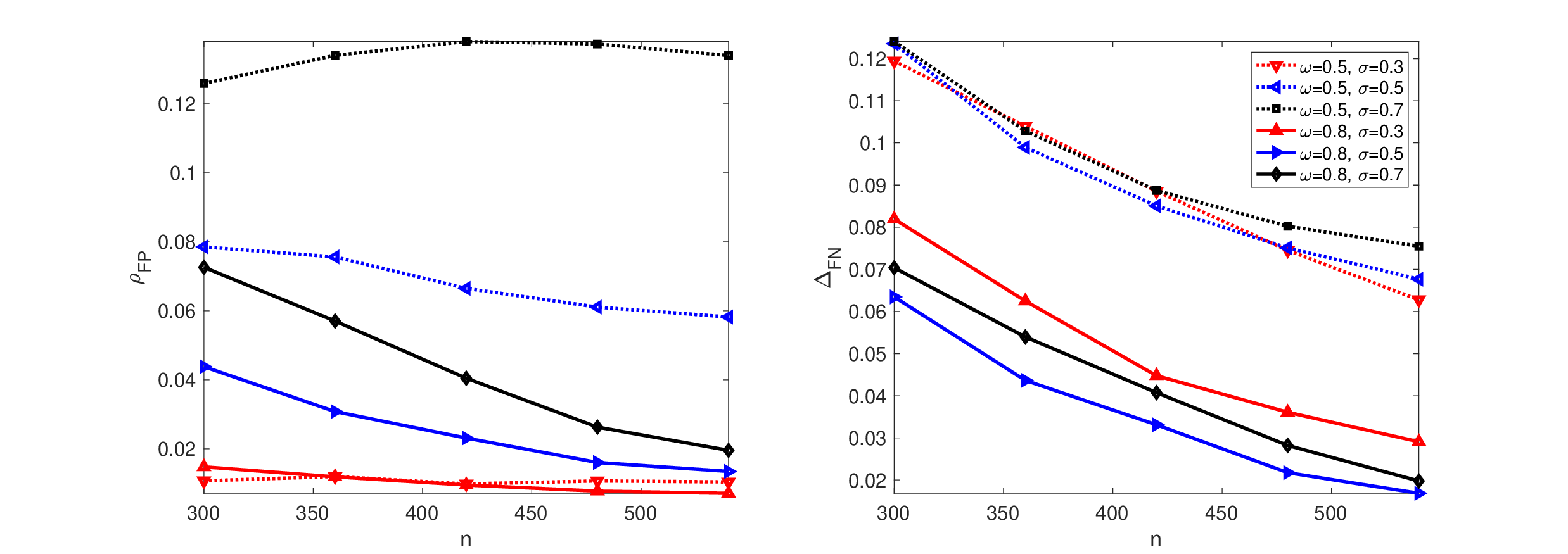}  }
\centerline{ \includegraphics[width = 17cm]{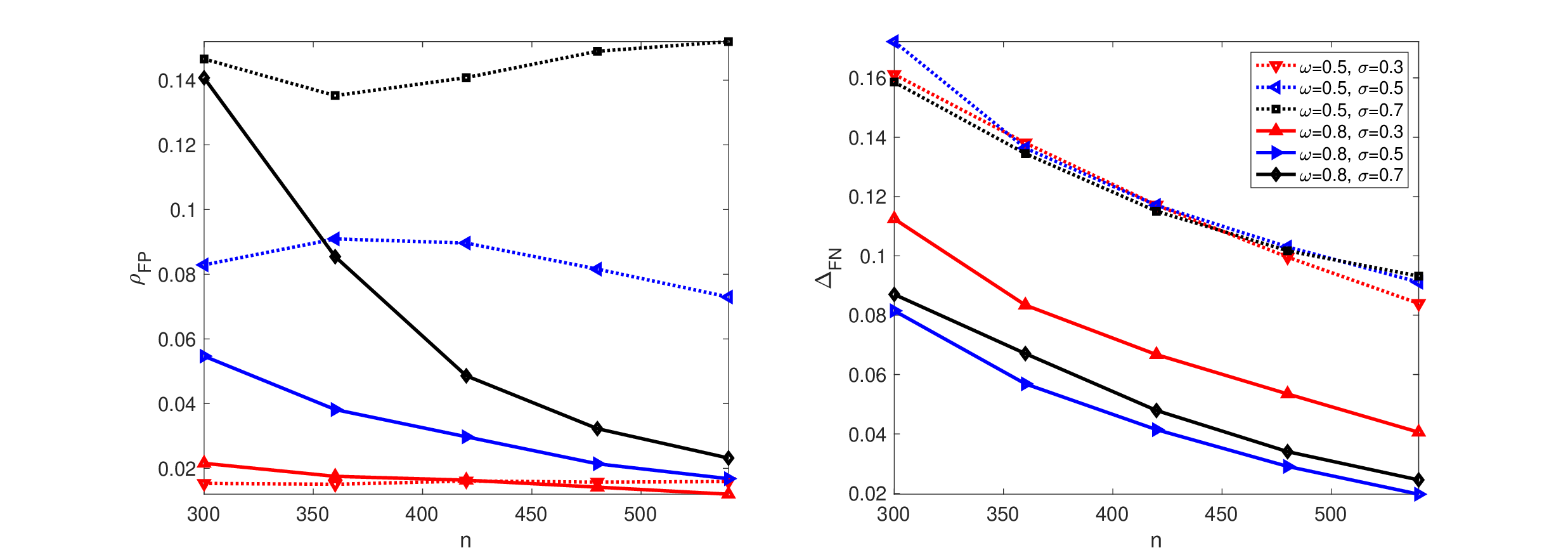}  }

\caption{ The false positive rates $\rho_{FP}$  (left panels)
and the rates $\Delta_{FN}$ (right panels) for $K=4$ (top), $K=5$ (middle) and 
$K=6$ (bottom) clusters. The rates are evaluated over 100 simulation runs. 
The number of nodes ranges from $n=300$ to $n=540$ with the increments of 60. 
Dashed lines represent the results for $\om=0.5$ and solid lines represent the results for $\om=0.8$;
$\sigma=0.3$  (red), $\sigma=0.5$ (blue) and $\sigma=0.7$ (black).
}
\label{mn:fig2}
\end{figure}

Now we use SSC to find the clustering matrix $\hat{Z}$. Since the diagonal elements of matrix $A$ are unavailable, 
we initially set $A_{ii}=0$, $i=1,...,n$, and solve optimization problem (\ref{mn:Elastic_Net})  with  
$\gamma_1=30\rho(A)$ and $\gamma_2=125(1-\rho(A))$, where  $\rho(A)$ is the density of matrix $A$,   
the proportion of  nonzero entries of $A$. The values of $\gamma_1$ and $\gamma_2$ 
have been obtained empirically by testing on synthetic networks.
After matrix $\widehat{W}$ of weights is  evaluated, we obtain the  clustering matrix
$\hat{Z}$ by applying spectral clustering to $|\widehat{W}| + |\widehat{W}^{T}|$, 
as it was described in Section~\ref{sec:clustering}. In this paper, we use the normalized 
cut algorithm \cite{shi2000normalized} 
to perform spectral clustering.
Given $\hat{Z}$, we generate matrix $A(\hat{Z}, K) = \scrP_{\hat{Z}, K}^T A \scrP_{\hat{Z}, K}$ with blocks $A^{(k,l)}(\hat{Z}, K)$,
$k,l=1,\ldots, K$, and obtain $\hat{\Theta}^{(k,l)}(\hat{Z}, K)$ by using the rank one approximation for each of the blocks. 
Finally, we  estimate  matrix $P$ by $\hat{P} =  \hat{P}(\hat{Z}, \hat{K})$ using  formula~\eqref{eq:P_total_est} with $\hat{K}=K$.


Figure   \ref{mn:fig1}  represents the accuracy of SSC in terms of 
the average estimation errors $n^{-2}\, \|{\hat{P}-P}\|_{F}^{2}$
and the average clustering errors $\Err (\hat{Z},Z)$ defined in \eqref{eq:misclustered} for $K=4, 5$ and 6, respectively, and 
the number of nodes ranging from $n=300$  to $n =540$ with the increments of 60.
 The left panels display the clustering errors $\Err (\hat{Z},Z)$ 
while the right ones exhibit the estimation errors $n^{-2}\, \|{\hat{P}-P}\|_{F}^{2}$,  as   functions of the number of nodes,
for two different values of the parameter $\omega$: $\omega =0.5$ (dashed lines) and 0.8 (solid lines) 
and three different values of the parameter $\sigma$: $\sigma =0.3$ (red lines), 0.5 (blue lines), and 0.7 (black lines).

Figure   \ref{mn:fig1}  shows  that sparsity has a  different affect on estimation and clustering errors. 
It is easy to see that as sparsity increases ($\sigma$ decreases), the estimation errors decrease.
On the other hand,  the difficulty of clustering depends on combination of the sparsity parameter $\sigma$ and  
the heterogeneity parameter $\omega$. Specifically, a denser network is easier to cluster when the network is more diverse 
(the heterogeneity parameter $\omega$ is larger), while for a very sparse network, heterogeneity of the network does not play much of a role.
Indeed, in all three graphs in the left half of Figure   \ref{mn:fig1}, the red curves, corresponding to the most sparse case ($\sigma =0.3$),
are close together while the black curves, corresponding to the least sparse case ($\sigma =0.7$), are further apart.
The graphs also show the effect of the number of clusters $K$ on the clustering errors. Indeed, for large $K$ ($K=6$), when $n$ is small ($n < 420$),
sparser network is not harder to cluster than denser one, perhaps because the diverse sparsity patterns make the network less uniform.
In summary, the difficulty of clustering depends on the interplay between sparsity and heterogeneity of the network.

Our procedure does not estimate the  set $J$ explicitly. Instead, we set $\hat{J} = \brJ = \bigcup_{k,l=1}^{K} \brJ_{k,l}$ 
where $\brJ_{k,l}$ is defined in \eqref{eq:brJ}. Our next objective is to evaluate how accurate $\brJ$ is, as an estimator of $J_*$. 
While  there are several  ways for doing this, below we use two measures, the false positive rate $\rho_{FP}$, defined as 
the proportion of zero entries in $P_*$ that are estimated by non-zeros in $\hat{P}$, and $\Delta_{FN}=\|P_*\|_{F}^{-1}\|X_*\|_{F}$, 
where $\|X_*\|_{F}$ is the Frobenius norm of nonzero entries in $P_*$ that are estimated by zeros in $\hat{P}$.
The reports on the accuracies of estimating $J_*$ are presented in Figure \ref{mn:fig2}. The left panels display $\rho_{FP}$ 
while the right ones exhibit $\Delta_{FN}$,  as   functions of the number of nodes for the same settings as in Figure  \ref{mn:fig1}.

The left panels of Figure \ref{mn:fig2} demonstrate  that the proportion of false positives $\rho_{FP}$
decreases as the network becomes more and more sparse and more heterogeneous (the proportions of false positives 
are smaller for smaller values of $\sig$ and larger values of $\om$). Again, the same as for Figure~\ref{mn:fig1},
the pattern emerges only when the number of nodes per community reaches some critical threshold. Indeed, as the bottom left panel 
of Figure \ref{mn:fig2} shows,  the  false positive rate is high, when the number of nodes is small.
The right hand side panels of Figure~\ref{mn:fig2} show that $\Delta_{FN}$, the relative norm of nonzero entries 
of $P_*$ estimated by zeros, is minimal for the moderately sparse network $\sig = 0.5$ and becomes smaller when the network is 
more heterogeneous. One can also notice that the values of  $\Delta_{FN}$ are almost independent of $\sig$ when the network
is relatively homogeneous ($\om =0.5$)  but become more diverse when the network becomes more diverse  ($\om =0.8$).



\begin{rem}\label{rem:Unknown_K}
{\bf Unknown number of clusters. }{\rm
In our previous simulations we treated  the true number of clusters as a known quantity. 
However, we can actually use $\hat{P}$ to obtain an   estimator $\hat{K}$  of $K$ 
by solving, for every suitable $K$, the   optimization problem  \eqref{eq:opt_for_K}, 
which can be equivalently rewritten as 
\be  \label{mn:Unknown_K}
\hat{K}=\mathop\text{argmin}_{K} \{ \|{\hat{P}-A}\|_{F}^{2} + \Pen  (n,J,K)\}.
\ee
The penalties  $\Pen (n,J,K)$ defined in \eqref{eq:penalties} are, however, motivated by 
the objective of setting it above the noise level with a very high probability. 
In our simulations, we also study the selection of an unknown $K$  using an empirical  version of this penalty 
\be \label{eq:smaller_pen}
\Pen(n,J,K)=\rho(A) n K \sqrt{\ln n\, (\ln K)^{3}}.
\ee
 
In order to assess the accuracy of  $\hat{K}$ as an estimator of $K$,  we evaluated 
$\hat{K}$ as a solution of optimization problem \eqref{mn:Unknown_K} with the penalty \eqref{eq:smaller_pen}
in each of the previous simulations settings over 100 simulation runs.
Table~\ref{mn:table1} in Section~\ref{sec:unknownK} of the Appendix  presents the   
relative frequencies of the estimators $\hat{K}$ of $K_*$ for $K_*$ 
ranging from 3 to 5,   $n=360$ and 480 and $\om=0.5$ and 0.8 and $\sigma = 0.4$, 0.6 and 0.8.
Table~\ref{mn:table1} confirms that for majority of settings, $\hat{K} = K_*$, i.e., the estimated and  the true number of clusters
coincide with high probability.

We would like to point out that the problem \eqref{mn:Elastic_Net} of finding weights   
is indeed strongly convex and it leads to a unique set of weights for every column of the 
adjacency matrix. However, the subsequent spectral clustering is not convex since it requires application of the
$K$-means clustering to the main $K$ eigenvectors of the weight matrix. 
The subspace clustering  is carried out   with a fixed number of clusters.
The number of clusters is then found as a solution of the discrete optimization problem \eqref{eq:opt_for_K}.
Therefore, even with the same adjacency matrix, due to random initialization of the $K$-means algorithm, 
the values of $\hat{K}$ may vary.
}
\end{rem}


\subsection{Real Data Examples}
\label{sec:real_data}

\begin{figure}[t]
\[ \includegraphics[height=6.0cm] {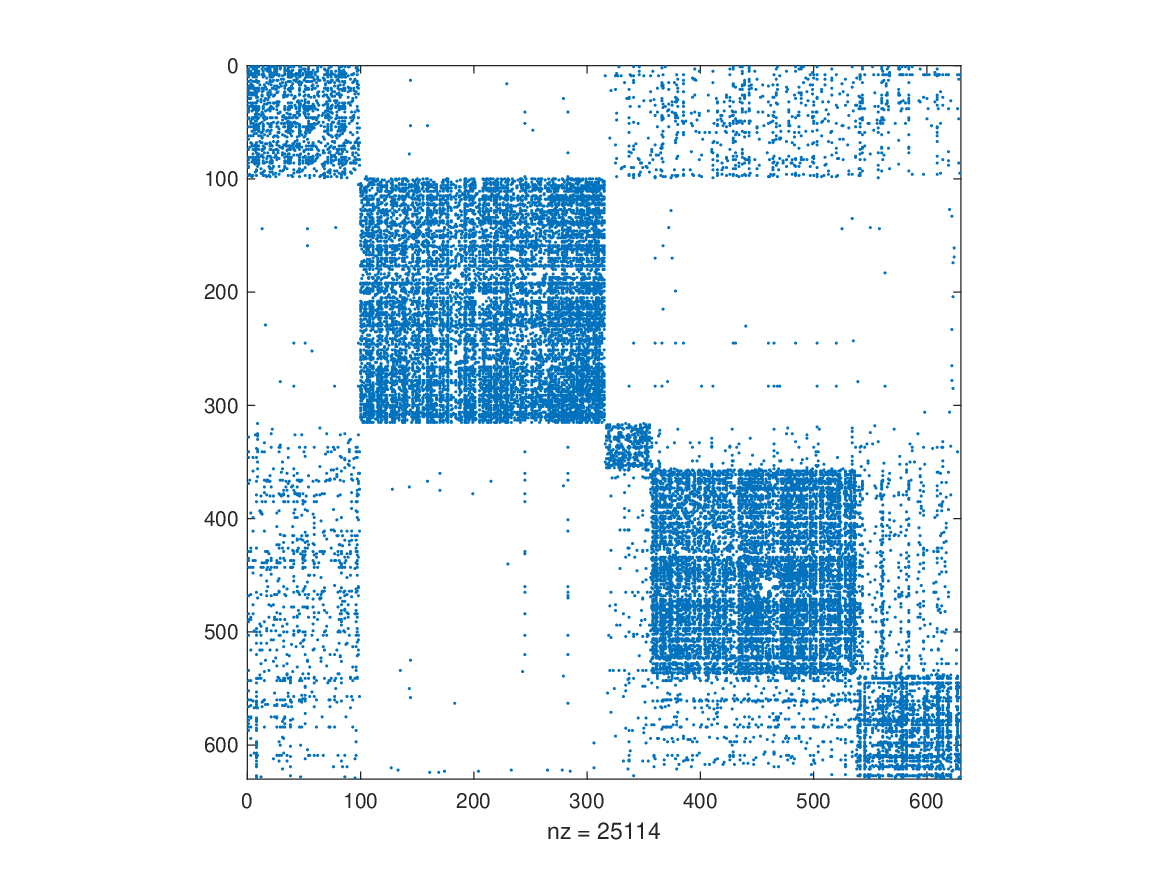} \hspace{5mm} 
\includegraphics[height=6.0cm]{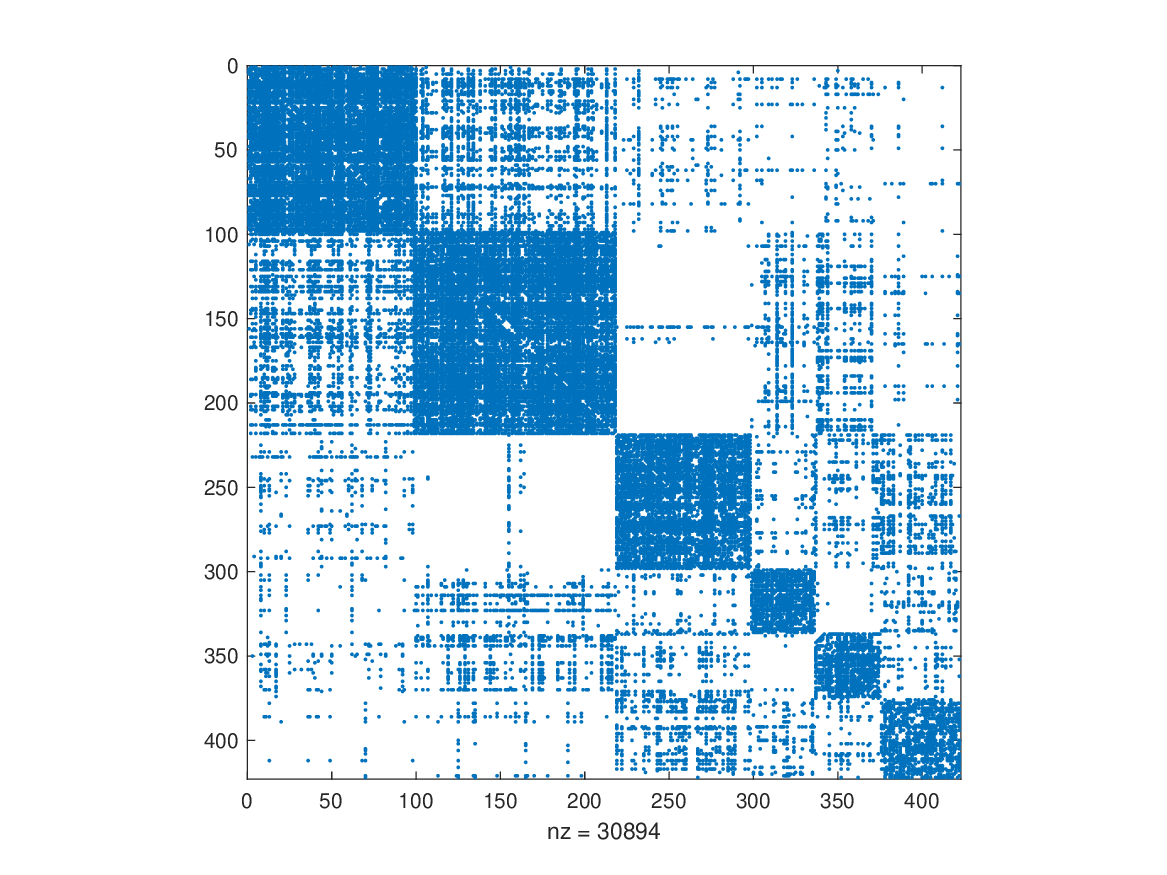}
\] 
  \caption{The adjacency matrices of the ego-network with 25114 nonzero 
entries and 5 clusters (left) and the brain network with 30894 nonzero entries and 6 clusters (right) after clustering  }
  \label{mn:fig_Adjacency matrices}
\end{figure}

In this section, we report the performance of SSC and our estimation procedure when they are applied to  two real life 
networks, an ego-network and a human brain network.

To study the ego-network, we use the dataset described comprehensively in \cite{leskovec2012learning}. 
An ego-network is a social network of a single person, with the exclusion of the person generating this network.
Users of social networking sites are usually provided with a tool that allows them to  organize their networks  into categories,
referred to, in \cite{leskovec2012learning}, as \textit{social circles}. Practically all major social networking cites 
provide such functionality, for example, ``circles'' on Google+, and ``lists'' on Facebook and Twitter.
Examples of such circles include   university classmates, sports team members, relatives, etc. Once circles are created by a  user,
they can be utilized, for example,   for content filtering (e.g. to filter status updates posted by distant acquaintances) or for privacy 
(e.g., to hide personal information from coworkers).

In this paper, we attempt to  recover social circles of an ego-network 
when only binary connection data is available. In particular, we formulate the problem of circle detection as a clustering problem 
on an individual ego-network. In principle,  circles can overlap or a circle can be a subset of another circle, hence, as an example in this paper, 
we study an ego-network with only few nodes overlap between the circles  which does not affect the performance of the clustering method. 
Specifically, we study an ego-network from Facebook where user profiles are treated as nodes 
and a friendship between two user profiles is considered as an edge between them. 
Since a friendship is a mutual tie, the ego-network is undirected.  
The ego-network studied in this paper, has 777 nodes with 17 circles, each circle containing between 2 to 225 nodes. For our study, 
we extract the five largest circles of the this network, obtaining a network with 629 nodes and 12557 edges. 
We carried out clustering of the nodes using the SSC and compared  the clustering assignments of SSC with the true class assignments.
The SSC provides 85\% accuracy. In addition,  we applied formula (\ref{mn:Unknown_K}) with $K$ ranging from 2 to 6 to 
the adjacency matrix with the randomly permuted rows (columns),  obtaining  the true number of clusters with 100\% accuracy over 100 runs. 
Figure \ref{mn:fig_Adjacency matrices} shows the adjacency matrix of the graph  after clustering (left),
which confirms that the network indeed follows the SPABM. Indeed, the SPABM is a very appropriate model for this example since
users display different degrees of connections to users in  other circles, and, furthermore, the network is sparse, which 
justifies the application of the SPABM.

  
Our second example involves   analyzing a human brain functional network, constructed on the basis of the resting-state functional MRI (rsfMRI). 
We use the the brain connectivity dataset presented as a GroupAverage rsfMRI matrix described in \cite{crossley2013cognitive}. 
In this dataset, the brain is partitioned into 638 distinct regions and a weighted graph is used to characterize the network topology. 
\cite{nicolini2017community} developed a new Asymptotical Surprise method, which is applied for clustering of the weighted graph.
Asymptotical Surprise detects 47 communities ranging from 1 to 133. Since the true clustering as well as the true number of clusters are unknown
for this dataset, we treat the results of the Asymptotical Surprise as the ground truth.

In order to generate a binary network, we set all nonzero weights to one in the GroupAverage rsfMRI matrix, 
obtaining a network with 18625 undirected edges.  For evaluating the performance of SSC 
on this network, we extract 6 largest communities derived by the Asymptotical Surprise, obtaining  a network with 422 nodes and 15447 edges. 
Applying \eqref{mn:Unknown_K}, with K ranging from 2 to 10, to the adjacency matrix with the randomly permuted rows (columns), 
we recovered the true number of clusters with 64\% accuracy over 100 simulation runs. For this true number of communities, our version of the SSC 
detects  the true communities with  94\% accuracy. Figure \ref{mn:fig_Adjacency matrices} (right) displays the adjacency matrix 
of the network after clustering, showing that the network is very sparse, thus, justifying  application of the SPABM to the data. 
  



\acks{The authors of the paper were  partially supported by National Science Foundation (NSF)  grants DMS-1712977 and DMS-2014928.  
We would also like to thank Drs.  Nicolini, Bordier and    Bifone for providing the brain dataset
together with the results of their clustering algorithm.}







\appendix
\section { }
\label{sec:appendA}
\setcounter{equation}{0} 

\renewcommand{\theequation}{A.\arabic{equation}}


\subsection{Accuracy of  Estimating the Number of Communities}
\label{sec:unknownK}

 Table~\ref{mn:table1} below 
presents the   relative frequencies of the estimators $\hat{K}$ of $K_*$ for $K_*$ 
ranging from 3 to 5,   $n=360$ and 480,   $\om=0.5$ and 0.8, and $\sigma = 0.4$, 0.6 and 0.8.
Table~\ref{mn:table1} confirms that for majority of settings, the estimated and the true number of clusters are equal,  $\hat{K} = K_*$, 
with high probability.


\begin{table} 
\begin{center}
\begin{tabular}{|c| c|c |c| c| c|c|c| }
\multicolumn{8}{ l }{  }\\
\hline \hline 
& & \multicolumn{6}{ |c |}{$n=360$} \\
\hline
& & \multicolumn{3}{ |c |}{$\omega=0.5$} & \multicolumn{3}{c|}{$\omega=0.8$}\\
\hline
$K_*$ & $\hat{K}$ & $\sigma=0.4$   & $\sigma=0.6$ & $\sigma=0.8$   & $\sigma=0.4$   & $\sigma=0.6$ & $\sigma=0.8$   \\
\cline{1-8} 
    & 2   & 0.01   & 0   & 0.01   & 0   & 0   &   0    \\  \cline{2-8}
    & 3   & \textbf{0.49}  & \textbf{0.62}   &  \textbf{0.62}  & \textbf{0.54}   &  \textbf{0.79}  & \textbf{0.76}     \\   \cline{2-8}     
 \textbf{3}  & 4   &  0.31  &  0.27  & 0.30   & 0.39   & 0.17   &   0.18    \\   \cline{2-8}
    & 5   & 0.15   & 0.09   &  0.06  &  0.06  & 0.04   &  0.06     \\     \cline{2-8}       
    & 6   & 0.04   & 0.02   &  0.01  & 0.01   & 0   &  0     \\      
\cline{2-8}    
\hline
\hline
%

\hline
 
    & 2   & 0   & 0   & 0   &  0  &  0  &  0     \\  \cline{2-8}
    & 3   & 0.01   & 0.01    & 0.05   & 0   &  0  & 0.01      \\   \cline{2-8}     
 \textbf{4}  & 4   & \textbf{0.66}   & \textbf{0.74}   &  \textbf{0.66}  & \textbf{0.72}   & \textbf{0.85}   & \textbf{0.81}      \\   \cline{2-8}
    & 5   & 0.22   & 0.22    &  0.25  &  0.23  & 0.15  &  0.16     \\     \cline{2-8}       
    & 6   & 0.11   &  0.03  & 0.04   & 0.05    & 0   & 0.02      \\      
\cline{2-8}      
\hline
\hline
%
 
    & 2   &  0  & 0   & 0.02   &  0  &  0  &  0     \\  \cline{2-8}
    & 3   &  0  &  0  & 0.03   & 0  & 0   & 0      \\   \cline{2-8}     
 \textbf{5}  & 4   & 0.05   & 0.07   & 0.23   & 0   &  0  &  0.08     \\   \cline{2-8}
    & 5   & \textbf{0.70}   &  \textbf{0.69}  &  \textbf{0.54}  &  \textbf{0.74}  & \textbf{0.84}  &  \textbf{0.84}     \\     \cline{2-8}       
    & 6   & 0.25   & 0.24   &  0.18  & 0.26   &  0.16  &  0.08     \\      
\cline{2-8}      
\hline
\hline

\multicolumn{8}{ l }{  }\\
\hline \hline 
& & \multicolumn{6}{ |c |}{$n=480$} \\
\hline
& & \multicolumn{3}{ |c |}{$\omega=0.5$} & \multicolumn{3}{c|}{$\omega=0.8$}\\
\hline
$K_*$ & $\hat{K}$ & $\sigma=0.4$   & $\sigma=0.6$ & $\sigma=0.8$   & $\sigma=0.4$   & $\sigma=0.6$ & $\sigma=0.8$   \\
\cline{1-8} 
    & 2   & 0   & 0   & 0   & 0   & 0   &   0    \\  \cline{2-8}
    & 3   & \textbf{0.64}  & \textbf{0.62}   &  \textbf{0.60}  & \textbf{0.54}   &  \textbf{0.73}  & \textbf{0.76}     \\   \cline{2-8}     
 \textbf{3}  & 4   &  0.26  &  0.17  & 0.31   & 0.32   & 0.24   &   0.19    \\   \cline{2-8}
    & 5   & 0.08   & 0.19   &  0.07  &  0.10  & 0.01   &  0.05     \\     \cline{2-8}       
    & 6   & 0.02   & 0.02   &  0.02  & 0.04   & 0.02   &  0     \\      
\cline{2-8}    
\hline
\hline
%

\hline
 
    & 2   & 0   & 0   & 0   &  0  &  0  &  0     \\  \cline{2-8}
    & 3   & 0.01   & 0    & 0   & 0   &  0  & 0      \\   \cline{2-8}     
 \textbf{4}  & 4   & \textbf{0.64}   & \textbf{0.68}   &  \textbf{0.76}  & \textbf{0.69}   & \textbf{0.74}   & \textbf{0.83}      \\   \cline{2-8}
    & 5   & 0.21   & 0.30    &  0.21  &  0.23  & 0.24  &  0.17     \\     \cline{2-8}       
    & 6   & 0.14   &  0.02  & 0.03   & 0.08    & 0.02   & 0      \\      
\cline{2-8}      
\hline
\hline
%
 
    & 2   &  0  & 0   & 0   &  0  &  0  &  0     \\  \cline{2-8}
    & 3   &  0  &  0  & 0.02   & 0  & 0   & 0      \\   \cline{2-8}     
 \textbf{5}  & 4   & 0.04   & 0.01   & 0.21   & 0   &  0  &  0.05     \\   \cline{2-8}
    & 5   & \textbf{0.65}   &  \textbf{0.78}  &  \textbf{0.65}  &  \textbf{0.77}  & \textbf{0.89}  &  \textbf{0.86}     \\     \cline{2-8}       
    & 6   & 0.31   & 0.21   &  0.12  & 0.23   &  0.11  &  0.09     \\      
\cline{2-8}      
\hline
\hline
\end{tabular} 
\end{center}
\caption{The relative frequencies of the estimators $\hat{K}$ of $K_*$ for $K_*$ ranging from 3 to 5,   
$n=360$ and 480 and $\om = 0.5$ and 0.8 and $\sigma = 0.4$, 0.6 and 0.8.} 
\label{mn:table1}
\end{table}


\subsection{Proof of Theorem~\ref {th:oracle}}
\label{sec:proof_t1}

{\bf Overview:} The proof follows the standard oracle inequality strategy. We bound the error $\norm{\hat{P} - P_*}_F^2$ 
by the random error term plus the difference between the values of the penalty function at  $K_* ,J_*$ and $\hat{K}$, $ \hat{J}$:
\bes
2 \Tr\left[(A - P_{*})^T (\hat{P} -P_{*})\right] + \Pen(n, {J_{*}}, K_{*}) -\Pen( n, {\hat{J}}, \hat{K}). 
\ees
Subsequently, we show that the random error term is bounded above by the sum of the $\Pen( n, {\hat{J}} , \hat{K})$ and 
a small multiple of $\norm{\hat{P} - P_*}_F^2$ with high probability. The latter leads to the conclusion that 
$\norm{\hat{P} - P_*}_F^2$ is smaller than a multiple of $\Pen(n, {J_{*}}, K_{*})$ with high probability. The details of the proof 
are given below.
\\

\noindent{Proof. }
Denote $\Xi = A - P_*$ and recall that,  given matrix $P_*$, entries  
$\Xi_{i,j} = A_{i,j}-(P_{*})_{ij}$ of $\Xi$  are the independent Bernoulli errors for $1 \leq i \leq j \leq n$ and $\Xi_{i,j}= \Xi_{j,i}$. 
 Then following notations \eqref{eq:permute}, for any  $Z$ and $K$
 \bes 
\Xi(Z,K) = \mathscr{P}_{Z,K}^T\Xi \mathscr{P}_{Z,K} \quad \mbox{and} \quad
 P_{*} (Z,K) = \mathscr{P}_{Z,K}^T P_{*} \mathscr{P}_{Z,K}.
\ees
Let $(\hat{\Theta}, \hat{Z},\hat{J},  {\hat{K}})$ be a solution of optimization problem  \eqref{eq:opt_main},
and  the estimator $\hat{P} \equiv \hat{P}(\hat{Z},\hat{J},\hat{K})$  of $P_*$ be of the form \eqref{eq:P_total_est}.
Since $A(Z,K)= \mathscr{P}_{Z,K}^T A \mathscr{P}_{Z,K} $, one has  $A =\mathscr{P}_{Z,K}A(Z,K) \mathscr{P}_{Z,K}^T $ and 
 it follows from \eqref{eq:opt_main} that
\begin{align*}
& \norm{\mathscr{P}_{\hat{Z},\hat{K}}^T A\mathscr{P}_{\hat{Z},\hat{K}}-\hat\Theta(\hat{Z},\hat{J},\hat{K}) }_F^2 + 
\Pen( n, {\hat{J}}, \hat{K}) \leq \\
&  \norm{\mathscr{P}_{Z_{*},K_{*}}^T A \mathscr{P}_{Z_{*},K_{*}}  - 
\mathscr{P}_{Z_{*},K_{*}}^T P_{*}  \mathscr{P}_{Z_{*},K_{*}}}_F^2  + \Pen(n, {J_{*}}, K_{*}) 
\end{align*}
Using orthogonality of permutation matrices, we can rewrite the previous inequality as
\be\label{eq:main_ineq0}
\norm{ A -\mathscr{P}_{\hat{Z},\hat{K}}\hat\Theta  {(\hat{Z},\hat{J},\hat{K})} 
\mathscr{P}_{\hat{Z},\hat{K}}^T }_F^2 \leq \norm{A - P_{*}}_F^2 + 
\Pen(n, {J_{*}}, K_{*}) -\Pen( n, {\hat{J}} , \hat{K})
\ee
Hence   \eqref{eq:main_ineq0} and  \eqref{eq:P_total_est} yield
\be\label{eq:main_ineq1}
 \norm{ A - \hat{P}}_F^2 \leq \norm{A - P_{*}}_F^2  +  \Pen(n, {J_{*}}, K_{*}) - \Pen( n, {\hat{J}} , \hat{K}) 
 \ee
Now adding and subtracting $P_{*}$ in the norm on the left side  of  \eqref{eq:main_ineq1}, we  rewrite \eqref{eq:main_ineq1} as
\be\label{eq:total:err}
\norm{\hat{P} -P_{*}}_F^2 \leq  \Delta {(\hat{Z},\hat{J},\hat{K})} + 
\Pen(n, {J_{*}}, K_{*}) - \Pen( n, {\hat{J}} , \hat{K})
\ee
where 
\bes\label{eq:DelZK}
 \Delta {(\hat{Z},\hat{J},\hat{K})} =  2 \Tr\left[\Xi^T (\hat{P} -P_{*})\right]. 
\ees
Again, using orthogonality of permutation matrices, we obtain
\bes
 \Delta {(\hat{Z},\hat{J},\hat{K})} =  2 \left \langle\Xi(\hat{Z},\hat{K}),(\hat\Theta {(\hat{Z},\hat{J},\hat{K})} 
- P_{*} (\hat{Z},\hat{K})) \right \rangle
\ees
where $\langle A,B\rangle = \Tr(A^TB)$.
Then, in the block form, $ \Delta {(\hat{Z},\hat{J},\hat{K})}$ appears as 
\be \label{eq:Del_blockform}
\Delta {(\hat{Z},\hat{J},\hat{K})} =  \displaystyle \sum_{k,l=1}^{\hat{K}} \Delta^{(k,l)} {(\hat{Z},\hat{J},\hat{K})}
\ee
with 
\bes 
\Delta^{(k,l)} {(\hat{Z},\hat{J},\hat{K})} = 2 \left\langle\Xi^{(k,l)}(\hat{Z},\hat{K}), 
\Pi_{\hat{u}, \hat{v}}\left(\Pi_{\hat{J}^{(k,l)}}  \left( A^{(k,l)}(\hat{Z},\hat{K})\right)\right) - 
P_{*}^{(k,l)} (\hat{Z},\hat{K}) \right\rangle.
\ees 
Here $\Pi_{\hat{u}, \hat{v}}$ is defined in \eqref{rank1_Proj}, and  $\hat {u}\equiv \hat {u}^{(k,l)} {(\hat{Z},\hat{J},\hat{K})}$ and 
$\hat {v} \equiv  \hat{v}^{(k,l)} {(\hat{Z},\hat{J},\hat{K})}$ 
are the singular vectors of\  $\pijhbAZh$  corresponding to the largest singular values of $\pijhbAZh$.
Let  $\tilde {u}=\tilde {u}^{(k,l)}({\hat{Z},\hat{J},\hat{K}})$ and $\tilde {v} =\tilde{v}^{(k,l)}({\hat{Z},\hat{J},\hat{K}})$ 
be the singular vectors of $\pijhbPsZh$ corresponding to the largest singular values of $\pijhbPsZh$, and
$\Pi_{\tilde{u}, \tilde{v}} ( \pijhbPsZh )$ be the rank one projection of $P_{*}^{(k,l)} (\hat{Z},\hat{K})$  defined in \eqref{rank1_Proj}.

We point out here that although all singular vectors depend on the block $(k,l)$,  as well as on $Z,J$ and $K$, we omit these dependences 
from the notations since, otherwise, the paper will become unreadable. In addition, vectors $\hat {u}$ and $\hat {v}$ have supports 
$\hat{J}_{k,l}$ and $\hat{J}_{l,k}$, respectively.
Recall that
\bes 
\Pi_{\hat{u},\hat{v}}  (\Pi_{\hat{J}^{(k,l)}}  (A^{(k,l)}({\hat{Z},\hat{K}} )))= \Pi_{\hat{u},\hat{v}} 
( \Pi_{\hat{J}^{(k,l)}}   (P_{*}^{(k,l)} (\hat{Z},\hat{K}) ) +\Pi_{\hat{J}^{(k,l)}}  ( \Xi^{(k,l)}(\hat{Z},\hat{K})))
\ees
Then, $\Delta^{(k,l)}(\hat{Z},\hat{J},\hat{K})$ can be partitioned into the sums of three components
\be \label{eq:Delkl_sum}
\Delta^{(k,l)}(\hat{Z},\hat{J},\hat{K}) = \Delta_1^{(k,l)}(\hat{Z},\hat{J},\hat{K}) + \Delta_2^{(k,l)}(\hat{Z},\hat{J},\hat{K}) + 
\Delta_3^{(k,l)}(\hat{Z},\hat{J},\hat{K}), \quad k,l = 1,2,\cdots, K,
\ee
where 
\begin{align}
\hspace*{-2.4cm}
\label{kl_blockdelta1_main_est_error}
&\Delta_1^{(k,l)}(\hat{Z},\hat{J},\hat{K})   =  2\, \left \langle \Xi^{(k,l)}(\hat{Z},\hat{K}), 
\Pi_{\hat{u},\hat{v}} (\Pi_{\hat{J}^{(k,l)}} (\Xi^{(k,l)}(\hat{Z},\hat{K})))\right \rangle \\
\label{kl_blockdelta2_main_est_error}
&\Delta_2^{(k,l)}(\hat{Z},\hat{J},\hat{K})  =   2\, \left\langle  \Xi^{(k,l)}(\hat{Z},\hat{K}), 
\Pi_{\tilde{u},\tilde{v}}(\pijhbPsZh)- P_{*}^{(k,l)} (\hat{Z},\hat{K})\right\rangle \\
\label{kl_blockdelta3_main_est_error}
&\Delta_3^{(k,l)}(\hat{Z},\hat{J},\hat{K})   =  2\left\langle \Xi^{(k,l)}(\hat{Z},\hat{K}), 
\Pi_{\hat{u},\hat{v}} (\pijhbPsZh )-\Pi_{\tilde{u},\tilde{v}}(\Pi_{\hat{J}^{(k,l)}} (P_{*}^{(k,l)} (\hat{Z},\hat{K})))\right\rangle    
\end{align}
With some abuse of notations,   for any matrix $B$ and any vectors $u,v$, 
let $\Pi_{u,v} \left( \Pi_{\hat{J}} (B(\hat{Z},\hat{K}))\right)$ be  the matrix 
with blocks $\Pi_{u,v} \left( \Pi_{\hat{J}^{(k,l)}}  (B^{(k,l)}(\hat{Z},\hat{K}))\right)$,  $k,l = 1,2,\cdots, \hat{K}$. 
Then, it follows from \eqref{eq:Delkl_sum}  that 
\be \label{eq:Del_sum}
\Delta(\hat{Z},\hat{J},\hat{K}) = \Delta_1(\hat{Z},\hat{J},\hat{K}) + \Delta_2(\hat{Z},\hat{J},\hat{K}) + \Delta_3(\hat{Z},\hat{J},\hat{K})  
\ee
where 
\beqn 
\label{delta1_main_est_error}
\Delta_1(\hat{Z},\hat{J},\hat{K}) &=&2\left \langle \Xi(\hat{Z},\hat{K}),\Pi_{\hat{u},\hat{v}} \left(   \pijhEZh \right) \right\rangle \\
\label{delta2_main_est_error}
\Delta_2(\hat{Z},\hat{J},\hat{K}) &=&  2\left\langle \Xi(\hat{Z},\hat{K}),\Pi_{\tilde{u},\tilde{v}}  \left( \pijhPsZh \right) - P_{*} (\hat{Z},\hat{K})\right\rangle\\
\label{delta3_main_est_error}
 \Delta_3(\hat{Z},\hat{J},\hat{K}) &= & 2\left\langle \Xi(\hat{Z},\hat{K}),\Pi_{\hat{u},\hat{v}} \left( \pijhPsZh \right)  - \Pi_{\tilde{u},\tilde{v}} \left( \pijhPsZh \right) \right\rangle.
\eeqn
Now, we need to derive an upper bound for each component in  \eqref{eq:Delkl_sum} and  \eqref{eq:Del_sum}.
\\


 
\noindent 
\underline{An upper bound for $\Delta_1(\hat{Z},\hat{J},\hat{K})$. }
Observe that 
\begin{align*}
|\Delta_1^{(k,l)}(\hat{Z},\hat{J},\hat{K})| 
& =  
2\, \norm{\Pi_{\hat{u},\hat{v}}  ( \Pi_{\hat{J}^{(k,l)}} (\Xi^{(k,l)}(\hat{Z},\hat{K})))}_{op}^2  
\leq 
2\, \norm{\Pi_{\hat{J}^{(k,l)}} ( \Xi^{(k,l)}(\hat{Z},\hat{K}) )}_{op}^2 
\end{align*}
Fix $t>0$ and let $\Omega_{1}$ be the set such that 
$\|\Pi_{\hat{J}}\left(\Xi (\hat{Z},\hat{K})\right)\|_{op}^2 \leq   F_1(n,\hat{J},\hat{K})  + C_2 t $.
According to Lemma~\ref{lem:prob_error_bound_main},  
\be \label{eq:P_Omega1}
\PP(\Omega_{1}) \geq  1-   \exp(-t),
\ee
and,  for $\omega \in \Omega_{1}$, one has 
\be \label{eq:Del1_bound}
|\Delta_1(\hat{Z},\hat{J},\hat{K})| \leq  
  2 \di\sum_{k,l = 1}^{\hat{K}} \|\Pi_{\hat{J}^{(k,l)}} (\Xi^{(k,l)}(\hat{Z},\hat{K}))\|_{op}^2 \leq 2\, F_1(n,\hat{J},\hat{K})  + 2\, C_2 t
\ee
where $F_1 (n, J, K)$ is defined by either \eqref{eq:F1}  or \eqref{eq:F1S} and $C_2$ is given in Lemma~\ref{lem:prob_bound_error}.
\\


\noindent 
\underline{An upper bound for $\Delta_2(\hat{Z},\hat{J},\hat{K})$. }
Now, consider $\Delta_2(\hat{Z},\hat{J},\hat{K})$ given by \eqref{delta2_main_est_error}.
Note that
\bes 
| \Delta_2(\hat{Z},\hat{J},\hat{K}) | = 2\, \|\Pi_{\tilde{u}, \tilde{v}}\left( \pijhPsZh \right)- 
P_{*} (\hat{Z},\hat{K}) \|_F \  |\langle \Xi(\hat{Z},\hat{K}),H_{\tilde{u}, \tilde{v}}(\hat{Z},\hat{J},\hat{K}) \rangle|,
\ees  
where 
\bes
H_{\tilde{u},\tilde{v}}(\hat{Z},\hat{J},\hat{K})= \frac{\Pi_{\tilde{u}, \tilde{v}}\left( \pijhPsZh \right)- 
P_{*} (\hat{Z},\hat{K})} {\|\Pi_{\tilde{u}, \tilde{v}}\left( \pijhPsZh \right)- 
P_{*} (\hat{Z},\hat{K})\|_F }
\ees
Since  for any $a,b$ and $\alpha_1 > 0$, one has $2ab \leq \alpha_1 a^2 +  b^2/\alpha_1$, obtain
\be \label{eq:Del2_sum} 
|\Delta_2(\hat{Z},\hat{J},\hat{K})| \leq  \alpha_1 \|\Pi_{\tilde{u}, \tilde{v}}\left( \pijhPsZh \right)- 
P_{*} (\hat{Z},\hat{K}) \|_F^2 +  \alpha_1^{-1}\ |\langle \Xi(\hat{Z},\hat{K}),H_{\tilde{u}, \tilde{v}}(\hat{Z},\hat{J},\hat{K}) \rangle|^2
\ee 
Observe that if $K, J$ and $Z \in \calM_{n,K}$ are fixed, then $H_{\tilde{u},\tilde{v}}(Z,J,K)$ 
is  fixed and, for any $K, J$ and $Z$, one has  $\|H_{\tilde{u},\tilde{v}}(Z,J,K)\|_F =1$.
Note also that, for fixed $K, J$ and $Z$,   matrix  $\Xi(Z,K) \in [0,1]^{n\times n}$ 
contains independent Bernoulli errors. It is well known that if $\xi$ is a vector of independent Bernoulli errors
and $h$ is any fixed vector with $\displaystyle \sum_{i =1}^n h_i^2 = 1$, then, for any $x>0$, the Hoeffding's inequality yields 
$$ \PP(|\xi^T h|^2> x) \leq 2 \exp(- x/2).$$
Since
$\langle \Xi(Z,K),H_{\tilde{u},\tilde{v}}(Z,J,K)\rangle =  [\vect(\Xi(Z,K))]^T  \vect(H_{\tilde{u},\tilde{v}}(Z,J,K))$,
applying the Hoeffding's inequality and accounting for the symmetry, we derive for any fixed $K,J$ and $Z$: 
\bes 
\PP  \left( | \langle \Xi(Z,K),H_{\tilde{u},\tilde{v}}(Z,J,K)\rangle|^2 - x > 0\right) \leq 2 \exp (-x/2). 
\ees
Hence, application of the union bound over $K, Z$ and $J$ leads to
\begin{align} \label{eq:Del2_union} 
& \  \PP  \left( | \langle \Xi(\hat{Z},\hat{K}),H_{\tilde{u},\tilde{v}}(\hat{Z},\hat{J},\hat{K}) \rangle|^2 - F_2(n,\hat{J} ,\hat{K}) > 2t\right) \\ 
\leq  
& \   \PP  \left( \underset{1 \leq K \leq n}{\max} \underset{J }{\max} \di\underset  {Z \in \mathcal{M}_{n,k}} {\max} 
[|\langle \Xi(Z,K),H_{\tilde{u},\tilde{v}}(Z,J,K) \rangle|^2 -F_2(n, J, K)]  > 2\, t \right) 
 \  \leq    2\, \exp(-t),  \nonumber
\end{align}
where $F_2(n,\hat{J},\hat{K})$   stands for $F_2^{(s)}(n,J,K)$ or  $F_2^{(ns)}(n,J,K)$ and 
\begin{align}
\hspace*{-2.4cm}
\label{eq:F2}
 F_2^{(ns)}(n,J,K) & =  2  \ln n  + 2 (n +2) \ln K + 2 |J| \ln (nKe/|J|)  \\
 \label{eq:F2S}
  F_2^{(s)}(n,J,K) & = 2 \, \di\sum_{k,l = 1}^K  |J_{k,l}| \ln (n_k e/|J_{k,l}|) 
+ 2 \left( \ln n + n \ln K + K \di\sum_{k=1}^K \ln n_k \right)
\end{align}
Using Lemma~\ref{lem:Pi_orth}, obtain that
\bes
\|\Pi_{\tilde{u}, \tilde{v}}\left( \pijhPsZh \right)- P_{*} (\hat{Z},\hat{K}) \|_F^2 \leq 
\|\Pi_{\hat{u}, \hat{v}}\left( \pijhPsZh \right)- P_{*} (\hat{Z},\hat{K}) \|_F^2 \leq 
\| \hat{P} - P_{*}  \|_F^2.
\ees 
Denote the set on which \eqref{eq:Del2_union} holds by $\Omega_{2}^c$, so that 
\be \label{eq:P_Omega2}
\PP(\Omega_{2}) \geq  1 - 2 \exp(-t).
\ee
Then inequalities \eqref{eq:Del2_sum}  and \eqref{eq:Del2_union} imply that, for any $\alpha_1>0$ and 
any $\omega \in \Omega_{2}$,  one has
\be \label{eq:Del2_bound}
 |\Delta_2(\hat{Z},\hat{J},\hat{K})|  \leq  \alpha_1  \| \hat{P} - P_{*}  \|_F^2 + 
\alpha_1^{-1} \, F_2(n,\hat{J},\hat{K}) + 2\, \alpha_1^{-1}\, t. 
\ee  
\\


\noindent 
\underline{An upper bound for $\Delta_3(\hat{Z},\hat{J},\hat{K})$. }
Now consider $\Delta_3(\hat{Z},\hat{J},\hat{K})$ defined in \eqref{delta3_main_est_error}
with components \eqref{kl_blockdelta3_main_est_error}.
Note that matrices 
$X_{k,l} = \Pi_{\hat{u},\hat{v}}  (\pijhbPsZh)-\Pi_{\tilde{u},\tilde{v}} (\Pi_{\hat{J}^{(k,l)}} (P_{*}^{(k,l)} (\hat{Z},\hat{K})))$ 
have ranks at most two.
Use the fact that (see, e.g., Giraud  (2014), page 123)
\be \label{eq:Ky-Fan}
\langle A, B \rangle  \leq \|A\|_{(2,r)} \|B\|_{(2,r)} \leq r\, \|A\|_{op} \|B\|_F,\quad r = \min \{\text{rank}(A), \text{rank}(B)\},
\ee
where, for any matrix $X$, $\|X\|_{(2,q)}$ is the Ky-Fan $(2,q)$ norm such that $\|X\|^2_{(2,q)} \leq \rank(X)\, \|X\|^2_{op}$. 
Applying inequality \eqref{eq:Ky-Fan} with $r=2$  to \eqref{kl_blockdelta3_main_est_error}, 
derive  that
\bes
|\Delta_3^{(k,l)}(\hat{Z},\hat{J},\hat{K})| \leq  4\,  \| \Pi_{\hat{J}^{(k,l)}} (\Xi^{(k,l)}(\hat{Z},\hat{K}))\|_{op}\
\left \|\Pi_{\hat{u},\hat{v}} (\pijhbPsZh)-\Pi_{\tilde{u},\tilde{v}} (\Pi_{\hat{J}^{(k,l)}} (P_{*}^{(k,l)} (\hat{Z},\hat{K}))) \right \|_F 
\ees
Then, for any $\alpha_2>0$, obtain 
\begin{align}  \label{eq:Del3_sum} 
& |\Delta_3 (\hat{Z},\hat{J},\hat{K}) | =  \di\sum_{k,l = 1}^{\hat{K}} |\Delta_3^{(k,l)}(\hat{Z},\hat{J},\hat{K})|  
  \leq \frac{2}{\alpha_2} \di\sum_{k,l = 1}^{\hat{K}} \| \Xi^{(k,l)}(\hat{Z},\hat{K} )\|_{op}^2  \\
& + 
2\, \alpha_2 \di\sum_{k,l = 1}^{\hat{K}}  
\|  \Pi_{\hat{u},\hat{v}} (\pijhbPsZh)-\Pi_{\tilde{u},\tilde{v}} (\Pi_{\hat{J}^{(k,l)}} (P_{*}^{(k,l)} (\hat{Z},\hat{K}))) \|_F^2  \nonumber
\end{align} 
Note that, by Lemma~\ref{lem:Pi_orth}, 
\begin{align*}
& \|\ \Pi_{\hat{u},\hat{v}}  ( \pijhbPsZh)-\Pi_{\tilde{u},\tilde{v}} (\Pi_{\hat{J}^{(k,l)}}  (P_{*}^{(k,l)} (\hat{Z},\hat{K})))\|_F^2  \leq  \nonumber \\
& 2\, \|\Pi_{\hat{u},\hat{v}} (\pijhbPsZh) - P_{*}^{(k,l)}(\hat{Z},\hat{K} )\|^2_F + 
2\, \|\Pi_{\tilde{u},\tilde{v}} (\Pi_{\hat{J}^{(k,l)}} (P_{*}^{(k,l)} (\hat{Z},\hat{K})))- P_{*}^{(k,l)}(\hat{Z},\hat{K} )\|^2_F \leq  \nonumber\\
%
%
& 4  \|\Pi_{\hat{u},\hat{v}} \left( \pijhbAZh \right) - P_{*}^{(k,l)}(\hat{Z},\hat{K} )\|^2_F 
= 4 \| \hat{\Theta}^{(k,l)}(\hat{Z},\hat{J},\hat{K})- P_{*}^{(k,l)}(\hat{Z},\hat{K} ) \|_F^2 
\end{align*}
 Therefore,
 \begin{align*}  
 &\di\sum_{k,l = 1}^{ \hat{K}}  \|   \Pi_{\hat{u},\hat{v}} \left( \pijhbPsZh \right)-\Pi_{\tilde{u},\tilde{v}}
 \left( \Pi_{\hat{J}^{(k,l)}}  \left(P_{*}^{(k,l)} (\hat{Z},\hat{K})\right) \right) \|_F^2 \leq  \nonumber\\
&  4  \norm{\hat{\Theta}(\hat{Z},\hat{J},\hat{K}) -P_{*}(\hat{Z},\hat{K} )}_F^2 
 = 4 \| \hat{P}  - P_{*}  \|_F^2   
 \end{align*}
Combining the last  inequality with  \eqref{eq:Del1_bound} and \eqref{eq:Del3_sum},  
obtain that     for any $\alpha_2>0$, $t>0$  and  $\omega \in \Omega_{1}$, one has
\be \label{eq:Del3_bound}
  |\Delta_3(\hat{Z},\hat{J},\hat{K})| \leq 8 \alpha_2 \| \hat{P}  - P_{*}  \|_F^2 + 
2\,\alpha_2^{-1}\, F_1(n,\hat{J},\hat{K})  + 2\,\alpha_2^{-1}\,  C_2\, t.
\ee
\\



\noindent 
\underline{An upper bound in probability. }
Let $\Omega =  \Omega_{1} \cap \Omega_{2}$. Then, \eqref{eq:P_Omega1} and \eqref{eq:P_Omega2}
imply that $\PP(\Omega) \geq 1 - 3 \exp(-t)$ and that, for $\om \in \Omega$,
inequalities \eqref{eq:Del1_bound}, \eqref{eq:Del2_bound} and \eqref{eq:Del3_bound} simultaneously   hold.
Hence,   \eqref{eq:Del_sum} implies  that, for any $\om \in \Omega$,
\bes 
|\Delta(\hat{Z},\hat{J},\hat{K})| \leq (2 + 2\,\alpha_2^{-1}) F_1(n,\hat{J},\hat{K})  + \alpha_1^{-1}\, F_2(n,\hat{J},\hat{K})
+ (\alpha_1 + 8 \alpha_2) \| \hat{P}  - P_{*}  \|_F^2 + 2\,(C_2 +  \alpha_1^{-1} + C_2\,\alpha_2^{-1})\, t.
\ees 
Combination of the last inequality and \eqref{eq:total:err} yields that, for $\alpha_1 +  8\alpha_2 <1$ and any $\om \in \Omega$,
\begin{align}  \label{est_error_const} 
 (1-\alpha_1 - 8\alpha_2)\, & \norm{\hat{P}  -P_{*}}_F^2 \leq   (2 + 2\,\alpha_2^{-1})\, F_1(n,\hat{J},\hat{K}) + \alpha_1^{-1}\, F_2(n,\hat{J},\hat{K})  \\
 & +\   \Pen(n,J_{*},K_{*}) - \Pen(n,\hat{J},\hat{K})   + 2\,(C_2 +  \alpha_1^{-1} + C_2\,\alpha_2^{-1})\, t.  \nonumber
\end{align}
Setting    
 $\Pen(n,\hat{J},\hat{K}) =  (2 + 2/\alpha_2) F_1(n,\hat{J},\hat{K})   + 1/\alpha_1  F_2(n,\hat{J},\hat{K})$,  
obtain the penalty as defined in \eqref{eq:penalties}--\eqref{eq:pen1}, with 
\be \label{eq:beta12}
 \beta_1 = \frac{2 (C_1 + C_2) (1 + \alpha_2)}{\alpha_2} + \frac{2}{\alpha_1}, \quad
\beta_2 = \frac{2 C_2  (1 + \alpha_2)}{\alpha_2} + \frac{2}{\alpha_1}.
 \ee 
 Dividing both sides of \eqref{est_error_const}   by $ (1-\alpha_1 - 8\alpha_2)$, obtain that 
\be \label{eq:tot_err_new}
\PP \lfi \|{\hat{P}  -P_{*}}\|_F^2 \leq   (1-\alpha_1 - 8\alpha_2)^{-1} \, \Pen(n,J_{*},K_{*}) + \tilde{C}\, t \rfi 
\geq 1 - 3 e^{-t}
\ee
where  $\tilde{C} = 2 (1-\alpha_1 - 8\alpha_2)^{-1}(C_2 + 1/\alpha_1 + C_2/\alpha_2)\, t $.
To obtain \eqref{eq:oracle} set $H_0 = (1-\alpha_1 - 8\alpha_2)^{-1}$.
\\


\noindent 
\underline{An upper bound in expectation. }
In order to obtain  the upper bound  \eqref{eq:oracle Expectation}  note that 
for $\xi = \|{\hat{P} -P_{*}}\|_F^2 -  H_0 \,  \Pen (n,K_{*})$, 
one has 
$\EE \|{\hat{P} - P_{*}}\|_F^2  = H_0 \,  \Pen (n,K_{*})  + \EE \xi$, 
where
\bes
\EE \xi \leq \int_0^{\infty} \PP(\xi > z) dz = \tilde{C}\int_0^{\infty} \PP(\xi >  \tilde{C}t) dt  
\leq \tilde{C}\int_0^{\infty} 3 \, e^{-t}\, dt  = 3\tilde{C},
\ees
which yields \eqref{eq:oracle Expectation}.\hfill\BlackBox


\subsection{Proof  of Theorem~\ref{th:clust}. }

Let $K$ be fixed, and  known, so that $K = K_*$ and, hence, $A (\hat{Z},K) \equiv A (\hat{Z})$  and so on.
Let $Z_*$ be the true clustering matrix and $J_*$ be the set of indices such that 
$ P_{i,j} (Z_*,K_* ) = 0 $ if $(i,j) \notin J_*$.  It follows from \eqref{eq:opt_ZK3} that \\
\bes
\begin{array}{ll}
\di\sum_{k,l = 1}^K \|A^{(k,l)}(\hat{Z}) -  \Pi_{(1)}   (\Pi_{\hat{J}^{(k,l)}} (A^{(k,l)}(\hat{Z}) ))\|_{F}^2  + \Pen(n,\hat{J}, K)\\
\leq   \di \sum_{k,l = 1}^K \|A^{(k,l)}(Z_{*}) -  \Pi_{(1)}   (\Pi_{J_*^{(k,l)}}  (A^{(k,l)}(Z_{*}) ) )\|_{F}^2  + \Pen(n,J_*, K)
\end{array}
\ees
where $\Pi_{(1)} (B)$ is the best rank one approximation of matrix $B$. 
Since for any $Z\in M_{n,K}$  and any $J$, one has 
\bes
\displaystyle \sum_{k,l = 1}^K \norm { A^{(k,l)}(Z) }_{F}^2  = \norm{A}_{F}^2, \ \  
\left\langle  A^{(k,l)}(Z),  \Pi_{(1)}  \lkr \Pi_{J^{(k,l)}} (A^{(k,l)}(Z))\rkr \right\rangle = 
\norm{ \Pi_{(1)}(\Pi_{J^{(k,l)}} (A^{(k,l)}(Z )))}_F^2
\ees
and $\Pen^{(1)} (n,K)$ does not depend on the sparsity set $J$, 
obtain from \eqref{eq:opt_ZK3},   with non-separable penalty  \eqref{eq:pen0ns}:
\begin{align}  \label{eq:clust1}  
\sum_{k,l = 1}^K \|\Pi_{(1)} \left(\Pi_{\hat{J}^{(k,l)}}\left(A^{(k,l)}(\hat{Z} )\right)\right)\|_{F}^2 & \geq 
  \sum_{k,l = 1}^K \|\Pi_{(1)}\left(\Pi_{J_{*}^{(k,l)}} \left(A^{(k,l)}(Z_{*} )\right)\right)\|_{F}^2 \\
& + \beta_1   |\hat{J}|  \ln (n K e/|\hat{J}|) - \beta_1   |J_* |  \ln (n K e/|J_*|). 
  \nonumber
\end{align}  
Denote, as before,   $ \Xi^{(k,l)}(Z) =  A^{(k,l)}(Z) -  P_{*}^{(k,l)}(Z) $. 
Note that, for any $J^{(k,l)}$, matrices $P_{*}^{(k,l)} (Z_{*})$ and $\Pi_{(1)}  (\Pi_{J^{(k,l)}} (A^{(k,l)}(Z)))$  have rank one, 
while for  $Z \neq Z_*$,  some $P_{*}^{(k,l)} (Z)$ may have ranks higher than one. 
Note that for any $Z\in \calM_{n,K}$  and any $J^{(k,l)}$ 
\begin{align}      \label{trian_ineq}
& \|  \Pi_{(1)}  (\Pi_{J^{(k,l)}} (A^{(k,l)}(Z)))\|_{F}    = \|  \Pi_{(1)}  (\Pi_{J^{(k,l)}} (A^{(k,l)}(Z)))\|_{op} \geq \\
& \|  P_{*}^{(k,l)} (Z)\|_{op} - \|  \Pi_{(1)}  (\Pi_{J^{(k,l)}} (\Xi^{(k,l)}(Z))   \|_{op} = 
\|  P_{*}^{(k,l)} (Z)\|_F - \|  \Pi_{(1)}  (\Pi_{J^{(k,l)}} (\Xi^{(k,l)}(Z))   \|_{op} \nonumber 
\end{align} 
Note that,  for $ (i,j) \notin J_{*}^{(k,l)}$, one has $\Xi_{i,j}^{(k,l)}(Z_*) =0$, since a Bernoulli 
random variable with zero mean is identically equal to zero. 
Therefore, for any set {$J^{(k,l)}$}, the matrix   $\Pi_{J^{(k,l)}}(\Xi^{(k,l)}(Z_{*}))$ has 
 $(J_{*})_{k,l} \cap J_{k,l}$ nonzero rows and $(J_{*})_{l,k} \cap J_{l,k}$  nonzero columns.
Thus, for any $t>0$, by Lemma~\ref{lem:prob_bound_error} 
\be \label{eq:ineq1}
\PP \left\{\sum_{k,l =1}^K  \norm{\left(\Pi_{J^{(k,l)}}(\Xi^{(k,l)}(Z_{*}))\right)}_{op}^2   
\leq  C_1 |J_{*} \cap J| + C_2\, t \right\}   \geq  1 - \exp (-t).
\ee 
Observe that, for any $a,b,c>0$, $a \geq b-c$ implies $b^2 \leq (1 + \tau)a^2 + (1 + 1/\tau)c^2$
for any $\tau>0$, so that $a^2 \geq b^2/(1+\tau) - c^2/\tau$.
Therefore, by \eqref{trian_ineq},  for any $\tau  \in (0,1)$, one has
\be\label{tau1_rel}
\| \Pi_{(1)} (\Pi_{J_{*}^{(k,l)}} (A^{(k,l)}(Z_{*}))) \|^2_{F} \geq 
(1 + \tau)^{-1}\, \|  P_{*}^{(k,l)} (Z)\|_F^2 - 
\tau^{-1}\, \|  \Pi_{(1)}  (\Pi_{J^{(k,l)}} (\Xi^{(k,l)}(Z))   \|_{op}^2.  
\ee
Hence, it follows from \eqref{eq:ineq1} and \eqref{tau1_rel}, that, for any  $\tau \in (0,1)$, any $t > 0$
\be  \label{eq:tau1_part}
\PP \left\{ \di \sum_{k,l =1}^K\,  \norm{\Pi_{(1)} [\Pi_{{J_*}^{(k,l)}} (A^{(k,l)}(Z_{*}))]}_{F}^2  \geq 
\frac{1}{1 + \tau}\,   \norm{{P_{*}}}_{F}^2 - \frac{C_1|J_*|}{\tau} - \frac{C_2  t}{\tau}  \right\} \geq 1 - e^{-t}.
\ee
On the other hand, for any $\tau_0 \in (0,1)$, derive
\bes 
 \norm{ \Pi_{(1)} [\Pi_{\hat{J}^{(k,l)}} (A^{(k,l)}(\hat{Z}))]}^2_{F} \leq
 (1 + \tau_0)  \norm{ \Pi_{\hat{J}^{(k,l)}} \lkr P_{*}^{(k,l)} (\hat{Z}) \rkr }_{op}^2 + 
 (1 + 1/\tau_0)    \norm{ \Pi_{\hat{J}^{(k,l)}} \Xi^{(k,l)} (\hat{Z})}_{op}^2.
\ees 
Taking a union bound similarly to Lemma~\ref{lem:prob_error_bound_main} and recalling that $K$ is fixed, 
obtain for any  $t > 0$
\bes  
\PP \lfi \di \sum_{k,l =1}^K   \norm{\Pi_{\hat{J}^{(k,l)}} (\Xi^{(k,l)}(\hat{Z})) }_{op}^2 \leq 
(C_1+ C_2) |\hat{J}| \ln (nKe/|\hat{J}|) + C_2 (2 \ln n + n \ln K + t)
 \rfi \geq  1- e^{-t}
\ees
Therefore, for any $\tau_0 \in (0,1)$  and any $t>0$, derive
\begin{align} \label{eq:tau2_part}
& \PP \lfi \di \sum_{k,l = 1}^K  \norm{\Pi_{(1)}  [\Pi_{\hat{J}^{(k,l)}} (A^{(k,l)}(\hat{Z}))]}_{F}^2  \right. 
  \leq 
(1 + \tau_0) { \di \sum_{k,l = 1}^K}  \norm{ \Pi_{\hat{J}^{(k,l)}} P_{*}^{(k,l)} (\hat{Z})}_{op}^2   \\
& + 
\left. (1 + 1/\tau_0) \lkv (C_1+ C_2) |\hat{J}| \ln (nKe/|\hat{J}|) + C_2 (2 \ln n + n \ln K + t)
 \rkv   \rfi \geq 1 - e^{-t}, \nonumber 
\end{align}
Combining  \eqref{eq:clust1},  \eqref{eq:tau1_part}  and \eqref{eq:tau2_part},  
derive that, for any $\tau, \tau_0 \in (0,1)$ and any $t>0$, one has
with probability at least $1 - 2e^{-t}$
\begin{align*}  
& (1 + \tau_0)   \sum_{k,l = 1}^K   \norm{ \Pi_{\hat{J}^{(k,l)}} P_{*}^{(k,l)} (\hat{Z})}_{op}^2
+ (1 + 1/\tau_0) \lkv (C_1+ C_2) |\hat{J}| \ln (nKe/|\hat{J}|) + C_2 (2 \ln n + n \ln K + t) \rkv \geq \\
& (1 + \tau)^{-1}\,    \norm{P_{*}}_{F}^2 - \tau^{-1}\, C_1|J_*|  - \tau^{-1}\, C_2  t  
+ \beta_1   |\hat{J}|  \ln (n K e/|\hat{J}|) - \beta_1   |J_* |  \ln (n K e/|J_*|). 
\end{align*}
Recall that, by Lemma~\ref{lem:hatJ}, $\hat{J} (\hat{Z}_K , K) \subseteq \brJ (\hat{Z}_K , K) \subseteq \brJ_* (\hat{Z}_K , K)$ 
and $\hat{J}^{k,l} (\hat{Z}_K , K) \subseteq   (\brJ_*)^{k,l} (\hat{Z}_K , K)$
for any $(k,l)$, so that 
\bes
\| \Pi_{\hat{J}^{(k,l)}} P_{*}^{(k,l)} (\hat{Z})\|_{op}^2 \leq \| \Pi_{(\brJ_*)^{(k,l)}} P_{*}^{(k,l)} (\hat{Z})\|_{op}^2 
= \| P_{*}^{(k,l)} (\hat{Z})\|_{op}^2.
\ees 
Then, combining similar terms and multiplying both sides by $(1 + \tau)$,  obtain for any $\tau, \tau_0 \in (0,1)$ and any $t>0$,  
with probability at least $1 - 2e^{-t}$ 
\begin{align*} 
&  \norm{P_{*}}_{F}^2  - (1 + \tau_0)(1 + \tau) \sum_{k,l = 1}^K   \norm{ P_{*}^{(k,l)} (\hat{Z})}_{op}^2 \leq 
(1 + \tau) [\tau_0^{-1} (1 + \tau_0) - \beta_1] |\hat{J}|  \ln (n K e/|\hat{J}|) +\\
&  \beta_1  |J_*|  \ln (n K e/|J_*|) +   (1 + \tau) |J_*|  [C_1\, \tau^{-1} + \beta_1 \ln (n K e/|J_*|) + \\
& C_2 (1 + \tau) (1 + \tau_0) \tau_0^{-1} (2 \ln n + n \ln K) + C_2   (1 + \tau_0) (1 + \tau^{-1} + \tau_0^{-1}) \, t.
\end{align*}
Set $t = n \ln K$.
Let $\tau = \tau_0$ and $(1 + \tau_0)(1 + \tau) = 1 + \alpha_n$. Then, $\tau^{-1} = \alpha^{-1} (1 + \sqrt{1 + \alpha_n})$, and hence
$\tau^{-1} (1 + \tau)^l \asymp \alpha^{-1}$ for $l=0,1,2$. Taking into account that,  by Lemma~\ref{lem:hatJ}, 
$|\hat{J} (\hat{Z})| \leq   |\brJ_* (\hat{Z})|$ and that function $f(x) = x \ln (nKe/x)$ is an increasing function of $x$,   
derive that for any $\alpha_n>0$ and $t>0$ and some absolute positive constants  $H_1$ and $H_2$, one has 
with probability at least $1 - 2 e^{-t}$
 \begin{align} 
& \| P_{*}\| _{F}^2 -   (1 + \alpha_n)\, \sum_{k,l = 1}^K \|  P_{*}^{(k,l)} (\hat{Z}) \|_{op}^2   \leq  
H_2 |J_*| \ln (nKe/|J_*|) +  \nonumber \\
&    H_1\, \alpha_n^{-1}\, \lkv  |\brJ_* (\hat{Z})| \ln (nKe/|\brJ_* (\hat{Z})|)    
  +     |J_*| + n \ln K  \rkr     \label{eq:main_clust1}
\end{align}
The proof is completed by comparison between \eqref{eq:main_clust1} and \eqref{eq:cond_check}, 
and by the contradiction argument.\hfill\BlackBox


\subsection{Proofs of  Lemmas 1, 2, 3 and 4}
\label{sec:lem_Proof}


\noindent{\bf Proof of Lemma \ref{lem:brJ_Jstar}. }
Note that index $j$ is incorrectly identified if $j \in J_{l,k}^* \cap (\brJ_{l,k})^c$  or  $j \in \brJ_{l,k} \cap (J_{l,k}^*)^c$.
Since Bernoulli variable with zero mean is always equal to zero, the second case is impossible.
Observe that for  any $(k,l)$, one has $P_*^{(k,l)} \equiv P_*^{(k,l)}(Z_*, K_*)$ and
\bes
\sum_{i=1}^{n_k}(P_*)_{ij}^{(k,l)} \geq  n_k \varpi (n, K) \geq \tilde{C_0}  n K^{-1} \, \varpi (n, K) \ \mbox{if}\  j \in J_{l,k}^*, 
\quad 
\sum_{i=1}^{n_k}(P_*)_{ij}^{(k,l)} = 0\ \mbox{if}\   j \in (J_{l,k}^*)^c
\ees
Therefore, for any $(k,l)$ and $j \in J_{l,k}^*$, by Hoeffding inequality, 
\begin{align*}
& \PP (j \in (\brJ_{l,k})^c) =   \PP \left( \displaystyle\sum_{i=1}^{n_k} A_{ij}^{(k,l)}(Z_*,K_*)=0  \right) 
 = \PP \left( \displaystyle\sum_{i=1}^{n_k}\left[ A_{ij}^{(k,l)}(Z_*,K_*) - (P_*)_{ij}^{(k,l)}\right] 
= - \displaystyle\sum_{i=1}^{n_k}(P_*)_{ij}^{(k,l)}  \right) \leq \\
&  \PP \left( \displaystyle\sum_{i=1}^{n_k}\left[ A_{ij}^{(k,l)}(Z_*,K_*) - (P_*)_{ij}^{(k,l)}\right] 
\leq - \tilde{C_0}  n K_*^{-1}\, \varpi (n, K_*) \right) 
\leq \exp \lfi  -2\tilde{C_0}^2 n K_*^{-2}\, \varpi^2 (n, K_*)  \rfi.
\end{align*}
Hence, applying the lower bound for $\varpi^2 (n, K_*)$ and the union bound, obtain 
\begin{align*}
& \PP(J_* (Z_*, K_*) \neq  \brJ (Z_*, K_*)) \leq \displaystyle\sum_{k,l=1}^{K} \PP  (j \in J_{l,k}^* \cap (\brJ_{l,k})^c ) \leq \\
& K_*^2 \exp \lfi  -2\tilde{C_0}^2 n K_*^{-2}\, \varpi^2 (n, K_*)  \rfi 
  \leq  K_*^2 n^{-1} e^{-t} \leq e^{-t}
\end{align*}
 which completes the proof. \hfill\BlackBox
\\

\medskip


\noindent{\bf Proof of Lemma~\ref{lem:hatJ}. }
Since $(P_*)_{i,j}=0$ implies $A_{i,j}=0$, one has  $\brJ_{k,l} (\hat{Z}_K , K) \subseteq (\brJ_*)_{k,l} (\hat{Z}_K , K)$
and $\brJ (\hat{Z}_K , K) \subseteq \brJ_* (\hat{Z}_K , K)$.

In order to prove the first inclusions in \eqref{eq:hatJbrJ}, consider the following two optimization problems
\begin{align}  
 \tilde{J} (\hat{Z}_K, K)  & \in   \underset{J}{\text{argmin}}  
\left\{\displaystyle \sum_{k,l = 1}^K \norm {A^{(k,l)}(\hat{Z}_K, K) -  \Pi_{(1)} 
\lkr \Pi_{J^{(k,l)}} (A^{(k,l)} (\hat{Z}_K,K)) \rkr}_{F}^2  + \Pen(n,J,K) \right\}  \label{eq:opt_withJ} \\
\ddot{J} (\hat{Z}_K, K)  & \in   \underset{J}{\text{argmin}}  
\left\{\displaystyle \sum_{k,l = 1}^K \norm {A^{(k,l)}(\hat{Z}_K, K) -  \Pi_{(1)} 
\lkr \Pi_{J^{(k,l)}} (A^{(k,l)} (\hat{Z}_K,K)) \rkr}_{F}^2   \right\}  \label{eq:opt_noJ}
\end{align}
Since  $\Pen (n,J,K)$ is an increasing function of $|J|$ (for a non-separable penalty) or of $|J_{k,l}|$
(for a  separable penalty), one has 
\be \label{eq:tilde_ddot}
 (\tilde{J})_{k,l}   (\hat{Z}_K , K) \subseteq (\ddot{J})_{k,l} (\hat{Z}_K , K), \quad
 \tilde{J}   (\hat{Z}_K , K) \subseteq \ddot{J}  (\hat{Z}_K , K)
\ee
On the other hand, one has $ \tilde{J} (\hat{Z}_K, K)  =  \hat{J} (\hat{Z}_K, K)$
since the right hand side of \eqref{eq:opt_withJ} is minimized at $\hat{J} (\hat{Z}_K, K)$.
In addition, it is easy to see that the right hand side of \eqref{eq:opt_noJ} takes the smallest value
at $\ddot{J} (\hat{Z}_K, K) = \brJ (\hat{Z}_K , K)$. Therefore,
\bes 
(\hat{J})_{k,l}   (\hat{Z}_K , K) \subseteq (\brJ)_{k,l} (\hat{Z}_K , K), \quad
 \hat{J}   (\hat{Z}_K , K) \subseteq \ddot{J}  (\brJ_K , K),
\ees
which completes the proof.
\hfill\BlackBox
 \\

\medskip 


\noindent{\bf Proof of Lemma \ref{lem:penalty}. }
Note that the difference between separable and non-separable penalty is   given by
\be  \label{eq:delns}
\Del^{n/s}   =   
  \Pen^{(ns)} (n,J,K) - \Pen^{(s)} (n,J,K) = 
\beta_1  \Del^{n/s}_1 + \beta_2 \Del^{n/s}_2   
\ee 
where
\begin{align*}
\Del^{n/s}_1 & =  |J|\, \ln \lkr \frac{n K e}{|J|} \rkr  - \sum_{k,l=1}^K\, |J_{k,l}|\, \ln \lkr \frac{n_k e}{|J_{k,l}|} \rkr, \quad
 \Del^{n/s}_2   =  2 \ln n - K\, \sum_{k =1}^K\, \ln n_k.
\end{align*}
Note that, due to the log-sum inequality (Theorem 17.1.2 of \cite{Cover:2006:EIT:1146355}),    $\Del^{n/s}_1  \leq 0$ with  $\Del^{n/s}_2  = 0$
if and only if $n_k/|J_{k,l}| = nK/|J|$ for every $k,l=1, \ldots, K$. In the extreme case where the nodes 
have nonzero connection probabilities only to the nodes 
in the same class, one has $|J_{k,l}|  = n_k$ for  $k=l$ and 0 otherwise, so that $|J| = n$. Then, 
$\Del^{n/s}_1 = n \ln K$, so that 
\be \label{eq:del1ns}
0 \leq \Del^{n/s}_1 \leq n \ln K.
\ee
Now, consider $\Del^{n/s}_2$. Note that application of the  log-sum inequality 
(Theorem 17.1.2 of \cite{Cover:2006:EIT:1146355})  yields
\bes
2 \ln n - K^2 \ln (n/K) \leq \Del^{n/s}_2 \leq 2 \ln n - K \ln (n+1 -K).
\ees
It is easy to see that $0 < K^2 \ln n \leq  n \ln K$ if $n \geq 8$ and $K \leq \sqrt{n/\ln n}$,
therefore,
\be \label{eq:del2ns}
2 \ln n - n \ln K  \leq \Del^{n/s}_2 \leq 2 \ln n.
\ee
Combining \eqref{eq:delns}--\eqref{eq:del2ns}, obtain that
\bes
\beta_2 (2 \ln n - n \ln K ) \leq \Del^{n/s}  \leq \beta_1 n \ln K + 2\,\beta_2 \ln n.
\ees
Hence, 
\begin{align*}
\Pen^{(ns)} (n,J,K) & \leq \Pen^{(s)} (n,J,K) + \beta_1 n \ln K + 2\,\beta_2 \ln n 
< (2 + \beta_1/\beta_2) \Pen^{(s)} (n,J,K) \\ 
 \Pen^{(s)} (n,J,K) & \leq \Pen^{(ns)} (n,J,K) + \beta_2 (2 \ln n - n \ln K )
< 2 \Pen^{(ns)} (n,J,K),  
\end{align*}
which leads to \eqref{eq:pen_rel}.\hfill\BlackBox

 \medskip 


\noindent{\bf Proof of Lemma \ref{lem:detect}. }
Note that 
$\Pi_{(1)} \lkr \Pi_{J_*^{(k,l)}} (P_*^{(k,l)}(Z_*)) \rkr  = \Pi_{(1)}  (P_*^{(k,l)}(Z_*)) = P_*^{(k,l)}(Z_*)$, so that 
the left hand side of inequality \eqref{eq:detect} is equal to identical zero. 
Also, $\Pi_{\brJ_*^{(k,l)}}(P_*^{(k,l)}(Z)) = P_*^{(k,l)}(Z)$, hence we need to prove that 
$\|P_*^{(k,l)}(Z) -  \Pi_{(1)}  (P_*^{(k,l)}(Z))   \|_{F} >0$ for at least one pair $(k,l)$,
$k,l=1, \ldots, K$.

Consider matrix  $Z \in \calM_{n, K_*}$ such that $Z$ cannot be obtained from $Z_*$ by
a permutation of columns. Let $i$ be a misclassified node, so that it belongs to communities $l_*$ and $l$
according to   $Z_*$ and $Z$, respectively.  Then, the  $i$-th column in the cluster $l_*$ of matrix $P_*$
is  vertical concatenation of vectors $\Lam^{(1,l_*)} *\Lam^{(l_*,1)}_i, \Lam^{(2,l_*)} *\Lam^{(l_*,2)}_i,
\ldots, \Lam^{(K,l_*)} *\Lam^{(l_*,K)}_i$. Since the node $i$ is connected to the network, 
there exists $t$ such that $\Lam^{(l_*,t)}_i >0$. When node $i$ is moved to cluster $l$, according to $Z$, 
the column $\Lam^{(t,l_*)} *\Lam^{(l_*,t)}_i$  is moved to the sub-matrix $P_*^{(t,l)}$ which contains multiples of vectors 
$\Lam^{(t,l_*)}$. 
Under Assumption~{\bf A1}, vectors ${\Lambda}^{(t,l)}$ and ${\Lambda}^{(t,l_*)}$ are linearly independent,
so that  the rank of sub-matrix $P_*^{(t,l)}(Z)$ is   at least two. Therefore,
$\|P_*^{(t,l)}(Z) -  \Pi_{(1)}  (P_*^{(t,l)}(Z))   \|_{F} >0$,
which completes the proof.


\subsection{Supplementary  Lemmas}
\label{sec:lem_suppl}


\begin{lem}  \label{lem:lowrank_approx}
Let $A$ and $B$ be arbitrary matrices in $\RR^{m \times n}$ and $u \in \RR^n$ and $v \in \RR^m$ be any unit vectors. 
Let   $\tilde{u}, \tilde{v}$ be the singular vectors of matrix $A$ corresponding to its largest singular value.
Then, 
\be \label{eq:Pi_uv}
\langle \Pi_{u,v}(B), A - \Pi_{u,v}(A) \rangle =0   \quad \mbox{and} \quad  
\|A - \Pi_{\tilde{u}, \tilde{v}}(A) \|  \leq  \|A - \Pi_{u,v}(A)\|,     
\ee 
so that,  the best rank one approximation of $A$ is given by $\Pi_{(1)}(A) = \Pi_{\tilde{u}, \tilde{v}}(A)$.
Here,  $\Pi_{u,v}(A)$ is defined in \eqref{rank1_Proj}. 
\end{lem}

\medskip


\begin{lem}  \label{lem:Pi_orth}
Let $A = P + \Xi$. Denote by $(\hat{u}, \hat{v})$ and $(u,v)$ the pairs of singular vectors of matrices $\Pi_{J} (A)$   
and $\Pi_{J} (P)$,  respectively, corresponding to their largest singular values. Then,
\be \label{eq:Pi_orth}
\|\Pi_{u,v} (\Pi_{J} (P) ) - P \|_F \leq \|\Pi_{\hat{u},\hat{v}} (\Pi_{J} (P) ) - P \|_F  \leq \|\Pi_{\hat{u},\hat{v}} (\Pi_{J} (A) ) - P \|_F
\ee 
where, for any matrix $X$,  $\Pi_{u,v} (X)$ is the projection of $X$
onto the pair of unit vectors $(u,v)$, given in \eqref{rank1_Proj},  and $\Pi_{J} (X)$ is the projection of the matrix $X$ 
onto the set of all matrices with the rectangular support  $J$.
\end{lem}


\noindent{\bf Proof.  }
Note that
\begin{align*}
& \|\Pi_{\hat{u},\hat{v}} (\Pi_{J} (A) ) - P \|_F^2  
 = \|\Pi_{\hat{u},\hat{v}} (\Pi_{J} (P + \Xi) ) - P \|_F^2 = \\
& \|\Pi_{\hat{u},\hat{v}} (\Pi_{J} (P)) +\Pi_{\hat{u},\hat{v}} (\Pi_{J}  (\Xi) ) - P \|_F^2 = \\
&  \| \Pi_{\hat{u},\hat{v}} (\Pi_{J}  (\Xi) ) + [\Pi_{\hat{u},\hat{v}} (\Pi_{J} (P))  - \Pi_{J} (P)] + [\Pi_{J} (P) - P] \|_F^2   
\end{align*}
Since matrices $\Pi_{\hat{u},\hat{v}} (\Pi_{J}(\Xi))$ and $[\Pi_{\hat{u},\hat{v}} (\Pi_{J} (P))  - \Pi_{J} (P)]$
are supported on the set of indices  $J$ and $\Pi_{J} (P) - P$ is supported on $J^c$, the latter matrix is orthogonal
to the first two.  On the other hand, 
$\Pi_{\hat{u},\hat{v}} (\Pi_{J}(\Xi))$ and $[\Pi_{\hat{u},\hat{v}} (\Pi_{J} (P))  - \Pi_{J} (P)] =  \Pi_{\hat{u},\hat{v}}^{\bot} (\Pi_{J} (P))$
are also orthogonal. 
Therefore,
\begin{align*} 
& \|\Pi_{\hat{u},\hat{v}} (\Pi_{J} (A) ) - P \|_F^2  
= \|\Pi_{\hat{u},\hat{v}} (\Pi_{J}  (\Xi) ) \|_F^2 +  \|\Pi_{\hat{u},\hat{v}} (\Pi_{J} (P))  - \Pi_{J} (P) \|_F^2 +  \| \Pi_{J} (P) - P \|_F^2 =\\
&  \|\Pi_{\hat{u},\hat{v}} (\Pi_{J}  (\Xi) ) \|_F^2 + \|\Pi_{\hat{u},\hat{v}} (\Pi_{J} (P))  - P \|_F^2   
\geq \|\Pi_{\hat{u},\hat{v}} (\Pi_{J} (P))  - P \|_F^2  
\geq \|\Pi_{u,v}(\Pi_{J} (P))  - P \|_F^2   
\end{align*}
where the last inequality follows from Lemma~\ref{lem:lowrank_approx}.\hfill\BlackBox
\\

\medskip


\begin{lem} \label{lem:prob_bound_error}
Let    elements of matrix $\Xi \in (-1,1)^{n \times n}$ be independent Bernoulli errors. 
Let matrix   $ \Xi$ be partitioned into $K^2$ sub-matrices $\Xi^{(k,l)}$ 
with supports $J^{(k,l)}  = J_{k,l}  \times J_{l,k}$,  $k,l = 1, \cdots, K$, such that
$\Xi^{(k,l)} = (\Xi^{(l,k)})^T$. 
Then, for any $x >0$  
\be \label{eq:lem3}
\PP \left\{ \sum_{k,l =1}^K \norm{ \Pi_{{J}^{(k,l)}} \left(\Xi^{(k,l)}\right)}_{op}^2  \leq   C_1|J| +  C_2 x \right\}   \geq  1 - \exp (-x),
\ee 
where $C_1$ and $C_2 $ are absolute constants independent of $n,K$ and sets $J_{k,l}$,  $k,l = 1, \cdots, K$.  
\end{lem}


\noindent{\bf Proof.} 
Denote $|J_{k,l}| = n_{k,l}$, $k,l = 1, \cdots, K$, and observe that matrices $\Xi^{(k,l)}$
are effectively of the size $n_{k,l} \times n_{l,k}$.
Consider $K(K+1)/2$-dimensional vectors $\xi$ and $\mu$ with elements $\xi_{k,l} = \|\Pi_{{J}^{(k,l)}}\left( \Xi^{(k,l)}\right)\|_{op}$ and 
$\mu_{k,l} =\EE \|\Pi_{{J}^{(k,l)}}\left(\Xi^{(k,l)}\right)\|_{op}$, $1 \leq k \leq l \leq  K$,   and let $\eta = \xi - \mu$.
Then, 
\be \label{norm_xi}
\Delta = \sum_{k,l =1}^K \norm{\Pi_{{J}^{(k,l)}} \left(\Xi^{(k,l)}\right)}_{op}^2   \leq  \| \xi \|^2 \leq  2 \| \eta \|^2 + 2 \| \mu \|^2
\ee 
Hence,  we need to construct the upper bounds for $\| \eta \|^2$ and $\| \mu \|^2$.

We start with constructing upper bounds for  $\| \mu \|^2$.
Let $\Xi_{i,j}^{(k,l)}$  be elements of the $(n_{k,l} \times n_{l,k})$-dimensional matrix $  \Pi_{{J}^{(k,l)}} \left(\Xi^{(k,l)}\right)$.
Then, $\EE(\Xi_{i,j}^{(k,l)}) = 0$ and, by Hoeffding's inequality,
$\EE \lfi \exp(\lambda \Xi_{i,j}^{(k,l)})\rfi \leq \exp\left( \lambda^2/8 \right)$.
Taking into account that Bernoulli errors are bounded by one in absolute value and applying 
Corollary 3.3 of  \cite{bandeira2016}  with $m = n_{k,l}$, $n = n_{l,k}$, $\sigma_{*}=1$, $\sigma_{1} = \sqrt{n_{l,k}}$ and 
$\sigma_2 = \sqrt{n_{k,l}}$, obtain
$$ 
\mu_{k,l}    \leq C_0\left(\sqrt{n_{k,l}} + \sqrt{n_{l,k}} + \sqrt{\ln(n_{k,l} \wedge n_{l,k})} \right)
$$
where $C_0$ is an absolute constant independent of $n_{k,l}$ and $n_{l,k}$.
Therefore, 
\be\label{NORM_MU} 
\|\mu\|^2 \leq 3C_0^2 \sum_{k,l =1}^K (n_{k,l} + n_{l,k}+ \ln(n_{k,l} \wedge n_{l,k})) 
\leq 6C_0^2 |J| + 3C_0^2  \sum_{k,l =1}^K  \ln(n_{k,l}).
\ee

Next, we show that, for any fixed partition,   $\eta_{k,l} = \xi_{k,l} - \mu_{k,l}$ are independent sub-gaussian random variables
when $1\leq k \leq  l \leq K$. Independence follows from the conditions of Lemma~\ref{lem:prob_bound_error}. To prove the 
sub-gaussian property, use Talagrand's concentration inequality (Theorem 6.10 of \cite{Boucheron2013}):
if $\Xi_1, \Xi_2,\Xi_3, \cdots , \Xi_n$ are independent random variables taking values in the interval $[0,1]$ and 
$f : [0,1]^n \rightarrow R$ is a separately convex function such that 
$|f(x) - f(y) | \leq \|x-y\|$  for  all  $x,y \in [0,1]^n$, 
then, for $Z = f(\Xi_1,\Xi_2,\Xi_3, \cdots , \Xi_n)$  and any $t > 0$, one has    
$\PP(Z > \EE Z +t) \leq \exp ( - t^2/2)$.  
Apply this theorem to vectors $\zeta_{k,l} = \vect (\Pi_{{J}^{(k,l)}} \left(\Xi^{(k,l)}\right))  \in [0,1]^{n_{k,l} \times n_{l,k}}$ and 
$f (\Pi_{{J}^{(k,l)}} \left(\Xi^{(k,l)}\right)) = f(\zeta_{k,l}) =  \norm{\Pi_{{J}^{(k,l)}} \left(\Xi^{(k,l)}\right)}_{op} $.
Note that, for any two matrices $\Xi$ and  $ \tilde{\Xi}$ of the same size, one has 
$\|\Xi - \tilde{\Xi} \|_{op}^2 \leq \|{\Xi - \tilde{\Xi}}\|_{F}^2  = \| \vect(\Xi) - \vect(\tilde{\Xi})\|^2$.
Then, applying Talagrand's inequality with  $Z = \|{\Pi_{{J}^{(k,l)}} \left(\Xi^{(k,l)}\right)}\|_{op}$ and $Z 
= -\|{\Pi_{{J}^{(k,l)}} \left(\Xi^{(k,l)}\right)}\|_{op}$,  obtain
\bes 
 \PP\left( \left| \|{\Pi_{{J}^{(k,l)}} \left(\Xi^{(k,l)}\right)}\|_{op} - \EE \|{\Pi_{{J}^{(k,l)}}\left(\Xi^{(k,l)}\right)}\|_{op} \right| > t \right) 
 \leq  2 \exp (-t^2/2).
\ees
Now, use the Lemma 5.5 of  \cite{vershynin_2012}  which states that the latter implies that, for any $t>0$ 
and some absolute constant $C_4>0$, 
\be \label{vershynin-app}
\EE\left[ \exp(t \eta_{k,l}) \right]  = \EE\left[ \exp(t(\xi_{k,l} - \mu_{k,l})) \right] \leq \exp (C_4 t^2/2).
\ee
Hence, $\eta_{k,l}$ are independent   sub-gaussian  random variables when $1\leq k \leq  l \leq K$.

In order to  obtain  an  upper bound for $\|{\eta}\|^2$, 
use Theorem 2.1 of \cite{Hsu2011}. Applying this theorem   with  $A =  I_{K(K+1)/2}$, $\mu =0$ and $\sigma^2 = C_4$ to a sub-vector  
$\tilde{\eta}$ of $\eta$ which contains components $\eta_{k,l}$ with $1\leq k \leq  l \leq K$,
obtain 
\bes
 \PP\left\{\|\tilde{\eta}\|^2  \geq C_4 \lkr K(K+1)/2   +  \sqrt{2\, K(K+1)\, x} + 2 x \rkr \rfi \leq   \exp (-x).
\ees
Since $\|{\eta}\|^2  \leq  2 \|{\tilde{\eta}}\|^2 $, derive 
\be\label{bound_eta}
 \PP\left\{\norm{\eta}^2  \geq 2C_4 K(K+1) + 6C_4x   \right\} \leq   \exp \left(-x \right)
\ee
Combination of formulas \eqref{norm_xi}  and \eqref{bound_eta} yield 
\bes
 \PP\left\{\norm{\xi}^2  \leq  2 \norm{\mu}^2  + 4C_4 K(K+1) + 12C_4x   \right\} \geq   1-  \exp \left(-x \right)
\ees
Plugging in $\norm{\mu}^2$  from \eqref{NORM_MU} into the last inequality, derive for any $x>0$ that
\be\label{main_res}
 \PP\left\{  \norm{\xi}^2  \leq  12C_0^2 |J| + 6C_0^2  \sum_{k,l =1}^K  \ln(n_{k,l}) + 4C_4 K(K+1) + 12C_4x   \right\} \geq   1-  \exp \left(-x \right).
\ee
Since $K(K+1) \leq  2K^2$ and 
$$6C_0^2  \di\sum_{k,l =1}^K  \ln(n_{k,l}) +  8C_4 K^2 \leq \max(6C_0^2, 8C_4)  \sum_{k,l =1}^K  \ln(n_{k,l} e)  \leq \max(6C_0^2, 8C_4)  |J|,$$ 
inequality \eqref{eq:lem3} holds with $C_1 = 12C_0^2 +  \max(6C_0^2, 8C_4)$  and  $C_2 = 12 C_4$. \hfill\BlackBox
\\

\medskip


\begin{lem}\label{lem:prob_error_bound_main}
For any $ t > 0$,
\be \label{rr:prob_bound_Xi_case1} 
\PP \left\{   \di\sum_{k,l = 1}^{\hat{K}} \norm{\Pi_{\hat{J}^{(k,l)}} \left(\Xi^{(k,l)}(\hat{Z},\hat{K})\right)}_{op}^2  -
F_1(n,\hat{J},\hat{K})  \leq   C_2 t \right\} \geq  1 - \exp{(-t)},
\ee
with $F_1(n,J,K) = F_1^{(ns)}(n,J,K)$ or $F_1(n,J,K)  = F_1^{(s)}(n,J,K)$, where
\begin{align}
\hspace*{-2.4cm}
\label{eq:F1}
& F_1^{(ns)}(n,J,K)   = (C_1+ C_2) |J| \ln (nKe/|J|) + C_2 (3 \ln n + n \ln K)\\
 \label{eq:F1S}
& F_1^{(s)}(n,J,K)   = (C_1+ C_2) \di\sum_{k,l = 1}^K  |J_{k,l}| \ln (n_k e/|J_{k,l}|)
+ C_2 \left( \ln n + n \ln K + K \di\sum_{k=1}^K \ln n_k \right)
\end{align}
and $C_1$ and $C_2$ are the absolute constants from Lemma~\ref{lem:prob_bound_error}.
\end{lem}


\noindent{\bf Proof. }
Note that $|J_{k,l}| \leq |J_{k,l}|  \ln (nKe/|J_{k,l}|)$, $|J| \leq |J|  \ln (nKe/|J|)$,
and also that $|J| =   \di\sum_{k,l = 1}^K |J_{k,l}|$.
First, let us prove the statement for $F_1(n,J,K) = F_1^{(ns)}(n,J,K)$. 
For this purpose, set $x = t + 3\ln n + n  \ln K  + |J|  \ln (nKe/|J|)$ 
 in Lemma \ref{lem:prob_bound_error} and apply the union bound over  $K \in [1,n]$,
$Z \in \calM_{n,K}$ and $J \subseteq  \{1, \ldots, nK\}$.
Obtain
\begin{align*}
&\PP \left\{   \di\sum_{k,l = 1}^{\hat{K}} \norm{\Pi_{\hat{J}^{(k,l)}} \left(\Xi^{(k,l)}(\hat{Z},\hat{K})\right)}_{op}^2  -  
  F_1^{(ns)}(n,\hat{J},\hat{K}) - C_2t  \geq 0  \right\} \\
%
%
%
\leq & \ 
\sum_{K=1}^n\   \sum_{Z\in \mathcal{M}_{n,K }} \ \sum_{j=1}^{nK}\  \sum_{|J|=j}\  
\PP \left\{  \di\sum_{k,l = 1}^K \|{ \Pi_{{J}^{(k,l)}} \left(\Xi^{(k,l)}(Z,K)\right)}\|_{op}^2 -  F_1^{(ns)}(n,J,K) 
\geq C_2 t\right\} \\ 
\leq & \ 
\sum_{K=1}^n\   \sum_{Z\in \mathcal{M}_{n,K }} \ \sum_{j=1}^{n K}\  \sum_{|J|=j}\  \exp(- t - 3\ln n - n  \ln K  - j  \ln (nKe/j))  \\
\leq & \ 
  \sum_{K=1}^n\   \sum_{j=1}^{nK}\  K^n {nK \choose j} \, \exp( - t - 3\ln n - n  \ln K  - j  \ln (nKe/j)) 
\leq \exp(-t).
\end{align*}


\noindent
In order to prove the statement for $F_1(n,J,K) = F_1^{(s)}(n,J,K)$, choose
\bes 
x = t +\ln n +n \ln K +   \di\sum_{k,l = 1}^K \left[ \ln (n_k) +  |J_{k,l}|  \ln (n_k \,e/|J_{k,l}|)\right]
\ees
in Lemma \ref{lem:prob_bound_error} and again  apply the union bound  over  $Z \in \calM_{n,K}$,  $K \in [1,n]$ and
$|J_{kl}| \in \{1, \ldots, n_k\}$, $k,l=1, \ldots, K$.  Obtain 
\begin{align*}
& \PP \left\{   \di\sum_{k,l = 1}^{\hat{K}} \norm{\Pi_{\hat{J}^{(k,l)}} \left(\Xi^{(k,l)}(\hat{Z},\hat{K})\right)}_{op}^2   - F_1^{(s)}(n,\hat{J},\hat{K}) - C_2t  \geq 0  \right\} \\
%
%
\leq & \ 
\sum_{K=1}^n\  \sum_{Z\in \mathcal{M}_{n,K }} \  \di\prod_{k,l = 1}^K\  \sum_{j_{k,l}=1}^{n_k} \ \sum_{|J_{k,l}| = j_{k,l}}\  
\PP \left\{  \di\sum_{k,l = 1}^K \|{ \Pi_{{J}^{(k,l)}} \left(\Xi^{(k,l)}(Z,K)\right)}\|_{op}^2 -  F_1^{(s)}(n,J,K) \geq C_2 t\right\}\\ 
\leq & \ 
\sum_{K=1}^n\  K^n \di\prod_{k,l = 1}^K\  \sum_{j_{k,l}=1}^{n_k} {n_k \choose j_{k,l}} \ 
\exp{\left(- t -\ln n -n \ln K -   \di\sum_{k,l = 1}^K \left[ \ln (n_k) +   j_{k,l}  \ln  (  n_k\, e/ j_{k,l}) \right]\right)} \\
 \leq & \ 
\exp{(-t)},
\end{align*}
which completes the proof.\hfill\BlackBox
\\

\medskip
 

\noindent{\bf Proof of the inequality \eqref{eq:cond_example}.\ }
For any $m$, denote $e_m = 1_m/\sqrt{m}$, so that $\|e_m\|=1$. 
Denote by    $\tlamkl$ and $\clamkl$ the portions of vectors $\Lam^{(k,l)}$, $k,l =1,2$, that, respectively,
stayed in the correct class and were moved to the wrong one by the erroneous clustering matrix $Z$. 
It is easy to check that, for $k=1,2$,  matrices 
$\tPkk \equiv P_{*}^{(k,k)} (Z)$  are $2 \times 2$--block matrices with blocks $\tlamkk (\tlamkk)^T$ and 
$\clamll (\clamll)^T$ on the main diagonal and $\clamlk (\tlamkl)^T$ and its transpose off the main diagonal.
Here,  for $k,l=1,2$, $k \neq l$, one has
\begin{align*} 
\tlamkk & = \sqrt{a  \, \tNk}\, e_{\tNk}, \quad 
\tlamkl = \sqrt{b\, |\tJk|}\, \lkr \sqrt{\tbek}\, e_{\tNk} + \sqrt{1 - \tbek}\, e_{\tNk}^\bot \rkr,\\
\clamkk & = \sqrt{a  \, \cNk}\, e_{\cNk}, \quad  
\clamkl = \sqrt{b\, |\cJk|} \lkr \sqrt{\cbek}\, e_{\cNk} + \sqrt{1 - \cbek}\, e_{\cNk}^\bot \rkr,
\end{align*}
where $e_m^\bot$ is a unit vector orthogonal to $e_m$. Consider matrices $U_k:\ (\tNk + \cNl) \times 4$, $k=1,2$, 
with the columns
\bes 
(U_k)_{:,1} = [e_{\tNk}; 0_{\cNl}], \ (U_k)_{:,2} =  [0_{\tNk}; e_{\cNl}], \ 
(U_k)_{:,3} = [e_{\tNk}^\bot; 0_{\cNl}], \ (U_k)_{:,4} = [0_{\tNk}^\bot; e_{\cNl}],
\ees
where $0_m$ is the $m$-dimensional zero column vector, and $[a;b]$ denotes 
the vector, obtained by stacking column vectors $a$ and $b$   together vertically.
Then, it is easy to verify that $U_k^T U_k = I_4$, and that  $\tPkk  = U_k H_k U_k^T$, where $H_k$ 
is the $4 \times 4$ symmetric matrix 
$$
H_k = [\tbk, R_k, 0, F_k; R_k, \cbl, G_k, 0; 0, G_k, 0, Q_k; F_k, 0, Q_k,  0]
$$
(with elements  listed row by row).  
Therefore, 
\be \label{eq:tildeP}
\|\tPkk\|_F^2  = \|H_k\|_F^2, \quad \|\tPkk\|_{op}^2  = \|H_k\|_{op}^2
\ee 
Consider the top left sub-matrix $\tHk = [\tbk, R_k; R_k, \tbl]$ of matrix $H_k$.
Let   $\lam_{1,k} \geq \lam_{2,k} \geq \lam_{3,k} \geq \lam_{4,k}$ and 
$\tlam_{1,k} \geq  \tlam_{2,k}$ be the eigenvalues   of matrices $H_k$ and $\tHk$, respectively.
Then, by Interlace Theorem (see \cite{Rao_Rao1998}, {\bf P 10.2.1})
with $m=4$ and $n=2$, obtain 
\be \label{eq:interlace}
\lam_{1,k} \geq \tlam_{1,k} \geq \lam_{3,k}, \quad 
\lam_{2,k} \geq \tlam_{2,k} \geq \lam_{4,k}
\ee
Observe that for any $\alpha >0$, one has 
\begin{align}  \label{eq:Hk_norm}
\|H_k\|_F^2 - (1 + \alpha) \|H_k\|_{op}^2 & \geq \lam_{2,k}^2 - \alpha \lam_{1,k}^2 \geq 
\lam_{2,k}^2 - \alpha \lkr \|H_k\|_F^2 -   \lam_{2,k}^2\rkr \\
&  = (1 + \alpha) \lam_{2,k}^2 - \alpha  \|H_k\|_F^2 
\geq (1 + \alpha) \tlam_{2,k}^2 - \alpha  \|H_k\|_F^2 \nonumber
\end{align}
Hence, by \eqref{eq:tildeP} and \eqref{eq:interlace}, for diagonal blocks,  derive 
\begin{align} \label{eq:DelD}
\Del_D & = \|\tPo\|^2_F + \|\tPt\|^2_F - (1+ \alpha_n)\, \lkr \|\tPo\|^2_{op}  + \|\tPt\|^2_{op}   \rkr \\
& \geq \lkr \tlam_{2,1}^2 + \tlam_{2,2}^2 \rkr - \alpha_n \, \lkr \|\tPo\|^2_F + \|\tPt\|^2_F  \rkr  \nonumber
\end{align}
Also, for non-diagonal blocks, one has 
\be \label{eq:DelND}
\Del_{ND} = 2 \|\tPot\|^2_F - 2 (1+ \alpha_n)\, \|\tPot\|^2_{op}
\geq - 2 \alpha_n \, \|\tPot\|^2_{op}  \geq - 2 \alpha_n \, \|\tPot\|^2_{F}
\ee
Combining \eqref{eq:DelD} and \eqref{eq:DelND}, obtain
\be \label{eq:cond0} 
\|P_{*}\|_{F}^2 -  (1 + \alpha_n)\,  \sum_{k,l = 1}^2 \|P_{*}^{(k,l)} (Z)\|_{op}^2 = 
\Del_D + \Del_{ND} \geq \tlam_{2,1}^2 + \tlam_{2,2}^2 - \alpha_n \, \|P\|^2_F
\ee
It is easy to check that 
\bes 
\tlam_{2,k}^2 = 1/4 \, \lkr \tbk + \cbl - \sqrt{ (\tbk + \cbl)^2 - 4 R_k^2} \rkr^2 
\geq (\tbk + \cbl)^{-2} \, (\tbk  \cbl - R_k^2)^2
\ees
Note that   $\tbk  \cbl - R_k^2 = \tbk  \cbl (1 - \rho_n^2\, \tbe_k^2\, \cbe_l^2)$. 
Also, due to $\max(\del_k, \del_l) \leq \del \leq 1/2 \leq 1 - \min(\del_k, \del_l)$, obtain
\bes 
\frac{\tbk  \cbl}{\tbk + \cbl} = \frac{ a N  (1 - \del_k) \del_l}{1 - \del_k + \del_l} \geq 
\frac{n\, a_n\, \del_l}{4}.  
\ees
Plugging the last two expressions into \eqref{eq:cond0} and taking into account that 
$\|P\|^2_F \leq  4 a^2 N^2 = a^2 n^2$, arrive at \eqref{eq:cond_example} with $\Del_n$ given by \eqref{eq:Deln}.
\\

\medskip



\vskip 0.2in
\bibliography{PABM}

\end{document}